\pdfoutput=1

\documentclass[11pt]{article}

\usepackage[final]{acl}
\usepackage{tabularx} 
\usepackage[table]{xcolor}
\definecolor{myblue}{rgb}{0.88, 0.94, 0.95} 

\usepackage{times}
\usepackage{latexsym}
\usepackage{graphicx}
\usepackage{subcaption}
\usepackage{xurl}
\usepackage{array}
\usepackage{booktabs}
\usepackage{makecell}

\usepackage{multirow}
\usepackage[T1]{fontenc}

\usepackage[utf8]{inputenc}

\usepackage{microtype}

\usepackage{inconsolata}

\usepackage{graphicx}

\title{MAviS: A Multimodal Conversational Assistant For Avian Species}

\author{
  \textbf{Yevheniia Kryklyvets},\; 
 \textbf{Mohammed Irfan Kurpath},\;   \textbf{Sahal Shaji Mullappilly},
\\
\textbf{Jinxing Zhou},\; 
\textbf{Fahad Shabzan Khan},\; 
\textbf{Rao Muhammad Anwer},
\\
\textbf{Salman Khan},\; \textbf{Hisham Cholakkal}
\\
 Mohamed bin Zayed
University of Artificial Intelligence
\\
 \small{
  \textbf{Correspondence:} 
  \{yevheniia.kryklyvets, jinxing.zhou,  hisham.cholakkal\}@mbzuai.ac.ae
 }
}

\makeatletter
\AtBeginDocument{
\let\@oldmaketitle\@maketitle
\renewcommand{\@maketitle}{\@oldmaketitle
  \vspace{-28pt}
  \includegraphics[width=\linewidth]{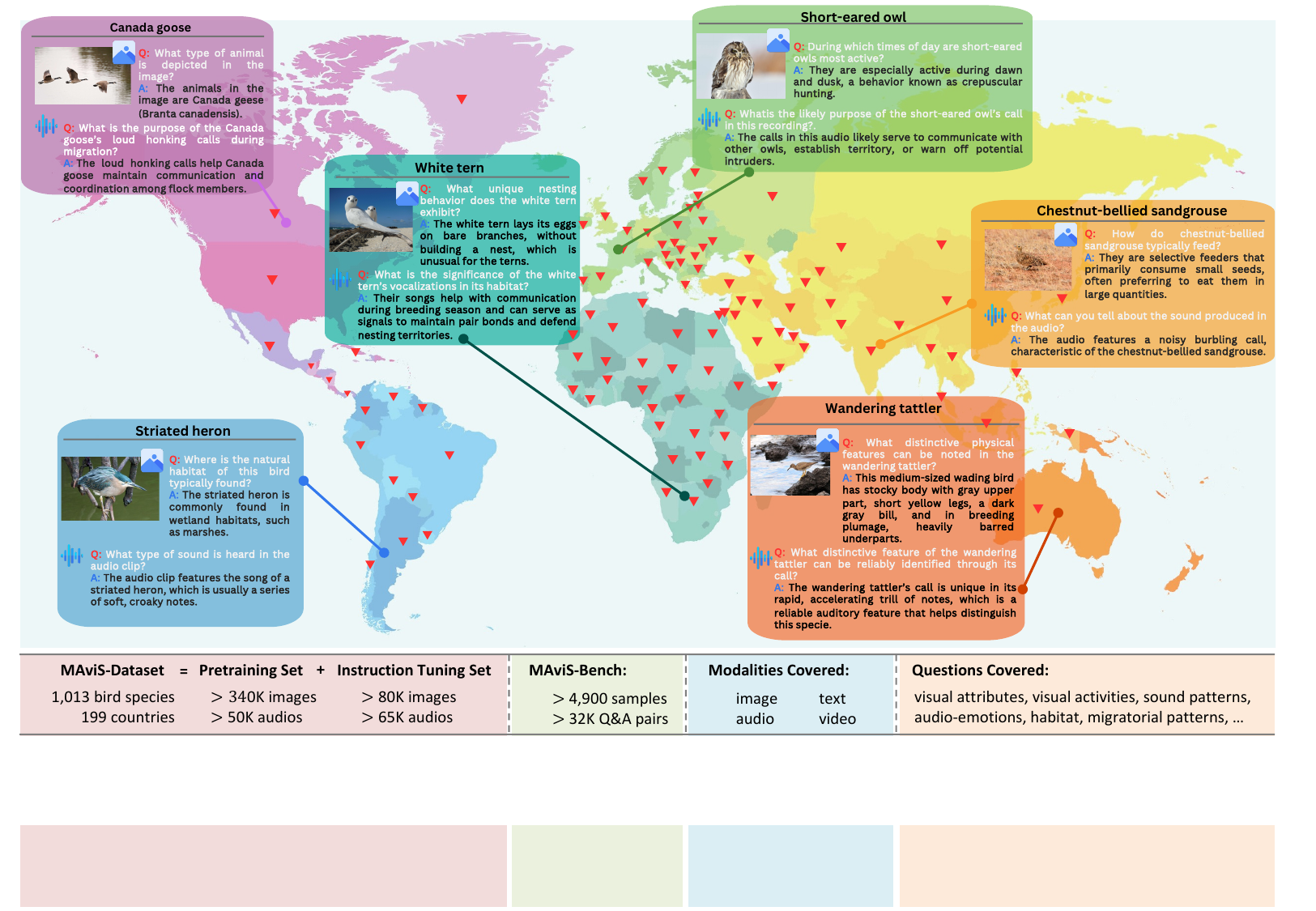}
  \vspace{-19pt}
  \captionof{figure}{\textbf{Overview of our MAviS suite.} The map illustrates the global distribution of over 1,000 bird species included in the proposed \textit{MAviS-Dataset}, spanning 199 countries, with geographic coverage indicated by red triangles. Six annotated Q\&A examples are displayed, each paired with image, audio, and text data, highlighting the multimodal and conversational nature. The bottom panel summarises key components of the dataset. MAviS-Dataset is organised into pretraining, instruction-tuning, and evaluation sets (\textit{MAviS-Bench}), covering diverse recognition and reasoning question types related to visual attributes, audio-based emotions, habitat and food habits, offering a valuable foundation for developing \textit{MAviS-Chat}, the proposed multimodal conversational assistant for avian species.
    }
    \label{fig:map}
  \vspace{10pt}
 }
}
\makeatother

\begin{document}

\maketitle

\begin{abstract}
Fine-grained understanding and species-specific, multimodal question answering are vital for advancing biodiversity conservation and ecological monitoring. However, existing multimodal large language models (MM-LLMs) face challenges when it comes to specialized topics like avian species, making it harder to provide accurate and contextually relevant information in these areas.
To address this limitation, we introduce the \textbf{MAviS-Dataset}, a large-scale multimodal avian species dataset 
that integrates image, audio, and text modalities for over 1,000 bird species, comprising both pretraining and instruction-tuning subsets enriched with structured question–answer pairs. 
Building on the MAviS-Dataset, we introduce \textbf{MAviS-Chat}, a multimodal LLM that supports audio, vision, and text designed for fine-grained species understanding, multimodal question answering, and scene-specific description generation. Finally, for quantitative evaluation, we present \textbf{MAviS-Bench}, a benchmark of over 25,000 Q\&A pairs designed to assess avian species-specific perceptual and reasoning abilities across modalities.
Experimental results show that MAviS-Chat outperforms the baseline MiniCPM-o-2.6 by a large margin, achieving state-of-the-art open-source results and demonstrating the effectiveness of our instruction-tuned MAviS-Dataset. Our findings highlight the necessity of domain-adaptive MM-LLMs for ecological applications. \textit{Our code, training data, evaluation benchmark, and models will be available at official repository.}
\end{abstract}

\section{Introduction}
\label{sec:intro}

Building accurate knowledge of species identity, geographic distribution, and evolutionary history is essential for sustainable development and ecosystem conservation~\cite{levis2024contributions,srivastava2005biodiversity}. However, basic information on many living organisms remains incomplete or inconsistent, making species identification a daunting task even for domain experts. 
Despite the number of applications for AI in conservation, from optimizing conservation planning and decision support to enhancing public engagement through AI-powered communication, the integration of these technologies into citizen science and conservation practices remains relatively young~\cite{ibrahim2024ai}. Challenges persist due to low-quality data from camera traps and drones, time-consuming and error-prone annotation processes, and the need for continuous retraining of models to accommodate evolving environmental conditions~\cite{conservation4040041}. At the same time, state-of-the-art generalized MM-LLMs suffer from inherent {limitations}: they are trained on broad datasets that lack domain-specific detail, 
struggle to distinguish fine-grained classes,
and are biased toward commonly occurring species~\cite{limits,zhang2024visually,lin2024towards}. As a result, these models are often ill-equipped to recognize rare or region-specific taxa accurately.


In this work, we explore the use of multimodal large models to support ecological intelligence, with a focus on fine-grained recognition in the avian domain. To this end, we propose \textbf{MAviS-Dataset}, a large-scale multimodal resource that spans \textbf{1,013 bird species} across the globe. It consolidates structured visual, acoustic, and textual information, supporting both model pretraining and downstream adaptation. 
Our MAviS-Dataset is constructed by repurposing publicly available image and audio datasets with species-level metadata (e.g., birdCLEF~\cite{birdclef-2021}, Tree of Life~\cite{treeoflife10m}, and iNaturalist~\cite{inaturalist2019}) by enriching them with newly curated annotations.
Our novelty lies in transforming these classificaion datasets into a domain-specific, instruction-tuned multimodal dataset. 
We introduce structured Q\&A pairs for over 1,000 bird species from 199 countries, covering multimodal reasoning tasks such as vocalization type classification, behavior-grounded interpretation, and ecological context inference (shown in Figure~\ref{fig:map}).
This restructuring creates the first large-scale instruction-aligned resource for wildlife multimodal LLMs, moving beyond simple species recognition toward fine-grained ecological reasoning.

MAviS-Dataset comprises two subsets. The \textit{pretraining} subset contains over \textbf{360,000} curated images and more than \textbf{55,000} audio clips, supporting broad feature representation learning. The \textit{fine-tuning} subset includes around \textbf{150,000} samples balanced across modalities for supervised instruction-tuning. To enable scalable and high-quality supervision, we develop an automatic annotation pipeline leveraging state-of-the-art audio/vision and language models, such as Qwen2-Audio~\cite{Qwen-Audio}, Llama-3.1~\cite{meta2024}, and Llama-3.2-Vision-Instruct~\cite{meta2024b}. As a result, the visual images or audio recordings are paired with detailed textual descriptions guided by structured prompts and external references. The fine-tuning examples are further enriched with instruction-based interactions, where each sample is paired with several question–answer pairs covering traits like appearance, sound, and habitat, providing targeted supervision for multimodal understanding. 

For evaluation, we introduce \textbf{MAviS-Bench}, a purpose-built benchmark comprising over \textbf{3,900} carefully selected samples and more than \textbf{25,000} instruction-response pairs. It includes both modality-specific and cross-modal tasks, assessing a model’s ability to classify species, interpret perceptual cues, and draw inferences from partial or implicit context.  
We also introduce a reference model, \textbf{MAviS-Chat}, which builds upon baseline Mini-CPM-o-2.6~\cite{yao2024minicpm} through  domain-specific adaptation with our dataset. We have evaluated our model on MAviS-Bench as well as on real-world videos comprising synchronized audio–visual signals. Our MAviS-Chat demonstrates strong performance in generating grounded, species-aware responses across multiple modalities, highlighting the benefits of our dataset for fine-grained avian species understanding in ecological contexts.

{In summary, our contributions are as follows:}  
\begin{enumerate}
    \item We construct MAviS-Dataset, the first large-scale multimodal resource dedicated to fine-grained avian species, integrating vision, audio, and text modalities across 1,013 bird species. 
    \item We establish MAviS-Bench, a multimodal benchmark designed to evaluate the fine-grained perceptual and reasoning capabilities of MM-LLMs in avian species understanding across audio, vision modalities. 

    \item {We present MAviS-Chat, a multimodal conversational assistant that supports audio and vision modalities and enables fine-grained, species-aware reasoning and responses across these modalities.} 
    \item We conduct extensive comparisons between MAviS-Chat and both commercial and state-of-the-art open-source MM-LLMs, demonstrating the effectiveness of our domain-adaptive pretraining and instruction tuning.

\end{enumerate}

\section{Related Work}
\label{sec:rel-work}

\subsection{Multimodal Large Language Models}

Recent advancements in MM-LLMs have led to significant progress in integrating data from diverse modalities, such as text, image, and audio.
By merging multiple sources of information, MM-LLMs can generate richer, context-aware outputs compared to unimodal models. 
These models have demonstrated state-of-the-art performance in applications that include image captioning~\cite{li2022blip,li2023blip,wang2022git}, visual question answering (VQ\&A)~\cite{liu2024improved,zhu2023minigpt,dai2023instructblip}, and understanding of audiovisual scenes~\cite{tang2024avicuna,zhang2023video,zhou2022avs,zhou2025avss,zhou2025think,guo2025aligned}. 

Despite their success in general tasks, MM-LLMs face significant challenges in fine-grained understanding of species, particularly in domains such as ornithology studied in this paper, where subtle morphological differences (e.g., feather patterns in birds) and acoustic signals (e.g., bird calls) must be accurately identified. These models struggle to capture those distinctions between species because they rely on generalized representations learned from large, often noisy datasets.

\subsection{Multimodal Large Language Models For Species Preservation}

AI models, particularly those that utilize deep learning, have been applied to various conservation tasks, such as identifying endangered species~\cite{atuhurra2024distilling}, monitoring ecosystem biodiversity~\cite{d2025mining}, and tracking wildlife populations~\cite{foyet_2024}.
However, it is noteworthy that the application of MM-LLMs remains underexplored.
{Most work in this field is limited to general species classification based on images or sounds, with little research focused on integrating these modalities into MM-LLMs, which can enable more comprehensive and context-aware species recognition.}
Moreover, most of the products in this field are proprietary, with limited data availability or model architectures for academic analysis. This makes it difficult to critically assess the methodologies, improve upon them, or adapt them for more specialized conservation use cases.

\subsection{Multimodal Datasets and Benchmarks}

Existing benchmarks for MM-LLM evaluation primarily focus on generalized tasks, and, as noted in~\cite{radford2021learningtransferablevisualmodels}, they often fail to challenge models with domain-specific tasks that require fine-grained distinctions.
Recent work, such as MMBench~\cite{liu2024mmbenchmultimodalmodelallaround}, introduces a strong evaluation framework for vision-language models, addressing issues such as ability assessment, quality control, and multilingual evaluation. 
A related effort in another specific domain is AgriBench~\cite{zhou2024agribenchhierarchicalagriculturebenchmark}, a benchmark designed to evaluate MM-LLM in agriculture.
However, none of them explicitly evaluates the fine-grained class discrimination capability of MM-LLMs.
Our work highlights this in avian species domain by creating specialized, large-scale, class-rich multimodal dataset along with its benchmark.

\section{MAviS-Dataset: The First Large-Scale Multimodal Avian Dataset}~\label{sec:dataset}
Generic models struggle to distinguish visually and acoustically similar species, limiting their effectiveness in real-world ecological applications. Addressing these challenges requires the development of a high-quality dataset. MAviS-Dataset is a large-scale multimodal dataset designed for pre-training and fine-tuning MM-LLMs in fine-grained bird species understanding.


\subsection{Data Sources and Collection}
Since the primary objective of MAviS-Dataset was to make it accessible to the research community, it was essential to identify data sources that could be freely leveraged. 
Moreover, multimodal datasets, especially those that incorporate audio, are significantly underrepresented compared to vision/video-based ones. Therefore, our first priority was to secure a comprehensive, high-quality audio dataset.

To address this, we utilized the BirdCLEF competition 2021-2024~\cite{birdclef-2021,birdclef-2022,birdclef-2023,birdclef-2024} datasets, providing professionally annotated bird species recordings. A key strength of BirdCLEF is its annually expanding geographic coverage, ensuring diverse regional representation. Combined datasets include 1,013 species from multiple continents, offering a rich and globally distributed collection of bird vocalizations. However, since BirdCLEF data inherently carry species imbalance - some species are recorded more frequently than others - we supplemented it with additional audio samples from iNaturalist~\cite{inaturalist2019}. This citizen science platform aggregates field recordings contributed by experts and enthusiasts alike. This addition helped balance underrepresented species and ensured more uniform coverage. However, the iNaturalist dataset was insufficient, particularly for rare species, with only $\sim$100 samples available for 28 species. To address this, we reached out to the Macaulay Library~\cite{library}, where we acquired over 3,000 high-quality bird recordings, filling in gaps for these underrepresented species.
It is important to note that due to strict access regulations, the Macaulay Library data will not be included in the released version of MAviS-Dataset. 
In summary, we collected more than 115,000 audio samples.

For image data, acquiring high-quality samples representing the 1,013 species included in the dataset was crucial. Here, BioCLIP\'s~\cite{stevens2024bioclip} and Tree of Life (ToL)~\cite{treeoflife10m} played a pivotal role. 
From there, we were able to retrieve visual data for 960 classes. To fill in the gaps and ensure a holistic representation of all classes, we also incorporated additional images from iNaturalist, applying a strict filtering criterion: only post-2021 data was used. This decision was critical because INat21~\cite{inat2021}, a large-scale dataset derived from iNaturalist and released in 2021, was already a component of ToL, and our objective was to avoid duplicates.
Finally, we obtain over 420,000 image samples.

In addition to images and audio, textual descriptions were curated for each species to enable multimodal learning. For each class, we created a data supplement object, which included a one-paragraph species description from eBird~\cite{ebird}, a widely trusted ornithological database, and a corpus of curated Wikipedia text, providing generalized ecological and taxonomic knowledge. This approach ensured that our dataset captured both expert-driven and community-sourced textual information, creating a well-rounded multimodal dataset for fine-tuned species recognition.

\subsection{Data Preprocessing}
Ensuring the quality and consistency of MAviS-Dataset required a preprocessing and annotation pipeline for each modality. Image, audio, and text data underwent distinct refinement procedures to enhance data integrity, reduce noise, and establish a solid foundation for multimodal learning.

All images were resized to a uniform resolution to standardize the dataset and underwent duplicate removal to prevent redundancy. Species labeling was carefully validated through cross-referencing annotations from iNaturalist and the Tree of Life database, ensuring taxonomic accuracy. Additionally, each image was supplemented with a one-paragraph species description, sourced from eBird and enriched with curated textual information from Wikipedia, providing ecological and morphological context to enhance multimodal learning.

Preprocessing of audio recordings included trimming, resampling, and noise reduction to enhance the quality and clarity of bird vocalizations. Given that raw bioacoustic recordings often contain ambient noise (e.g., wind, insect sounds, and anthropogenic disturbances), advanced filtering techniques were employed to isolate salient frequency ranges corresponding to bird vocalizations. This step was critical in minimizing species misclassification due to background interference. Furthermore, the dataset was carefully curated to ensure a balanced representation across species, addressing recording availability and quality disparities.

\subsection{Annotation for Pretraining Set}
\begin{figure}[t]
    \centering
    \includegraphics[width=0.48\textwidth]{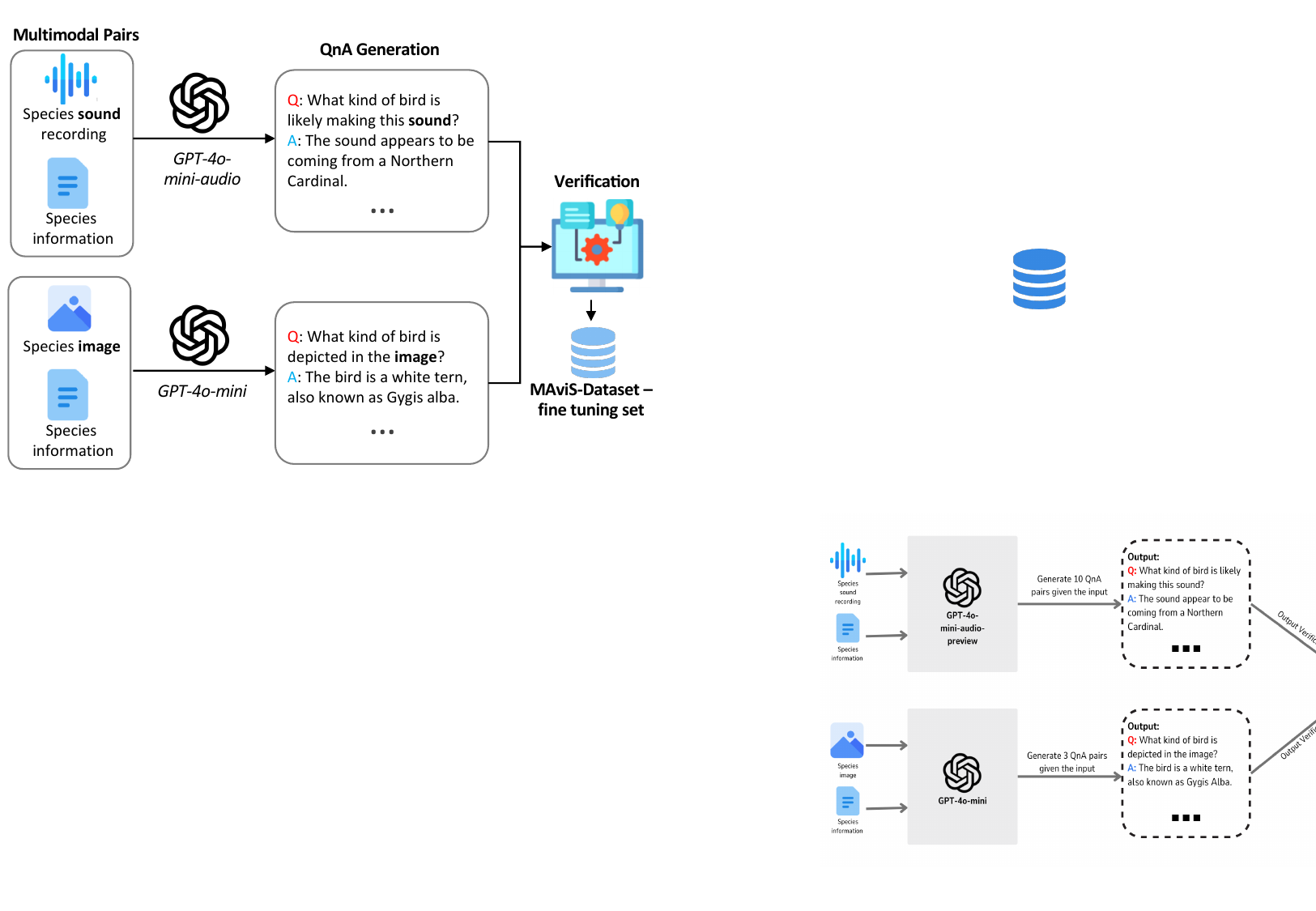}
    \vspace{-4ex}
    \caption{\textbf{Pipeline for generating image-text and audio-text annotations in the pretraining dataset.} The process combines large-scale public resources with structured AI-assisted enrichment to ensure semantic accuracy and species-specific grounding.}
    \vspace{-1ex}
    \label{fig:pip}
\end{figure}

We constructed a large-scale pretraining corpus using publicly available bird datasets, combining \textbf{audio} from birdCLEF 2021/22/23/24 and iNaturalist ($\sim$115k samples) and \textbf{images} from Tree of Life and iNaturalist ($\sim$400k samples). Since metadata of the original datasets was often limited to species ID, we enriched the data through a two-step, modality-specific annotation pipeline (Figure~\ref{fig:pip}). This process ensured accurate alignment and consistency for multimodal learning. As illustrated in Fig.~\ref{fig:map}, the dataset includes diverse resasoning-based Q\&A types beyond species recognition, including behavioral, acoustic, and ecological inference tasks.

\paragraph{Image-Text Annotations} 
Each image was first matched with species-level descriptions from sources such as eBird and Wikipedia. These were then expanded into descriptive paragraphs using the LLaMA 3.1-8B model~\cite{meta2024}, focusing on species-specific morphological traits. To further enhance quality, we employed LLaMA 3.2-11B-Vision-Instruct~\cite{meta2024b} with structured system prompts, generating species-aware, image-grounded annotations that emphasized fine-grained traits such as plumage, beak shape, posture, and other distinguishing features. This multimodal reasoning ensured high-fidelity image-text associations.

\paragraph{Audio-Text Annotations} 
For audio-text pairings, the first stage used Qwen2Audio~\cite{Qwen-Audio} to extract key acoustic features (e.g., pitch, rhythm, modulation), producing structured descriptions of bird vocalizations. In the second stage, these outputs were refined using LLaMA 3.1-8B~\cite{meta2024}, which integrated taxonomy-based reference information and behavioral context to enhance semantic richness and reduce hallucinations. This two-step strategy balanced annotation quality with computational efficiency, making it scalable to $\sim$115k samples.

\subsection{Fine-Tuning Set Curation}\label{sec:fine_tuning_set}
Following the annotation phase, we curated a uniformly distributed instruction-tuning subset from the enriched corpus: approximately 65,000 audio samples and 83,000 image samples. To enable instruction tuning, each sample was augmented with structured question–answer (Q\&A) pairs generated by GPT-4o-mini~\cite{openai2024gpt4ocard}. Prompts are detailed in Appendix~\ref{sec:appendixA}.

\begin{figure}[t]
    \centering
    \includegraphics[width=0.48\textwidth]{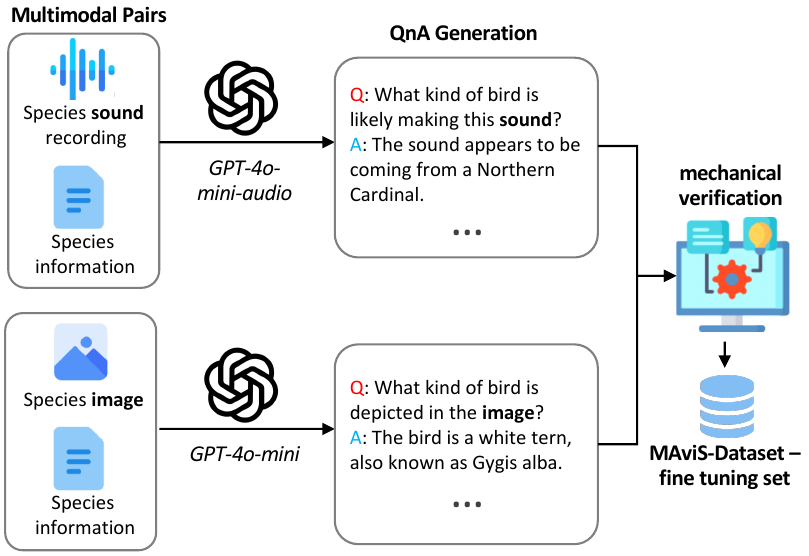}  
    \vspace{-4ex}
    
    \caption
    {\textbf{The curation process for the fine-tuning dataset}, detailing the sources, annotation, and refinement steps to ensure high-quality alignment. Broader multimodal Q\&A types are shown in Figure~\ref{fig:map} and detailed in Section~\ref{sec:fine_tuning_set}.}
    \vspace{-1ex}
    
    \label{fig:fine}
\end{figure}

Specifically, each \textbf{image} received up to 10 Q\&A pairs covering species identification, morphology, behavior, and habitat; while each \textbf{audio} sample received up to 3 Q\&A pairs focusing on sound classification, behavioral inference, and comparisons with similar-sounding species. Prompts were tailored to emphasize species-specific accuracy and fine-grained ecological reasoning. The resulting comprehensive instruction set forms the basis for fine-tuning, ensuring that models trained on it can effectively interpret multimodal data and generate accurate responses in downstream classification and retrieval tasks. Figure~\ref{fig:fine} illustrates the fine-tuning curation pipeline.

\subsection{Statistical Insights}

\begin{figure}[t]
    \centering
    \begin{subfigure}[b]{0.24\textwidth}
        \centering
        \includegraphics[width=\textwidth]{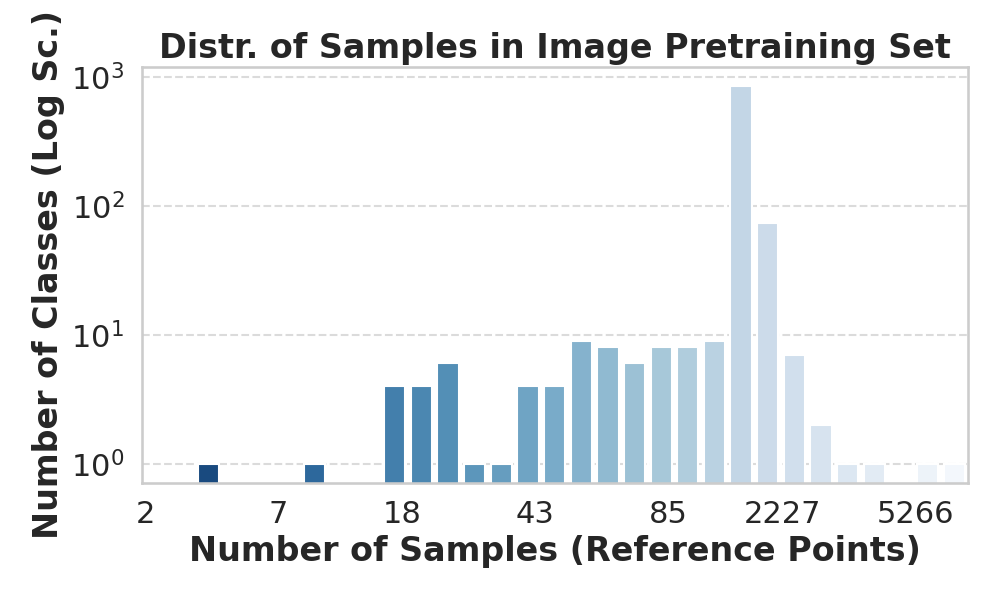}
        \caption*{ \textbf{(a)} Distribution of image samples per class in the pretraining set, covering over 360K images. 
        }
    \end{subfigure}
    \hfill
    \begin{subfigure}[b]{0.23\textwidth}
        \centering
        \includegraphics[width=\textwidth]{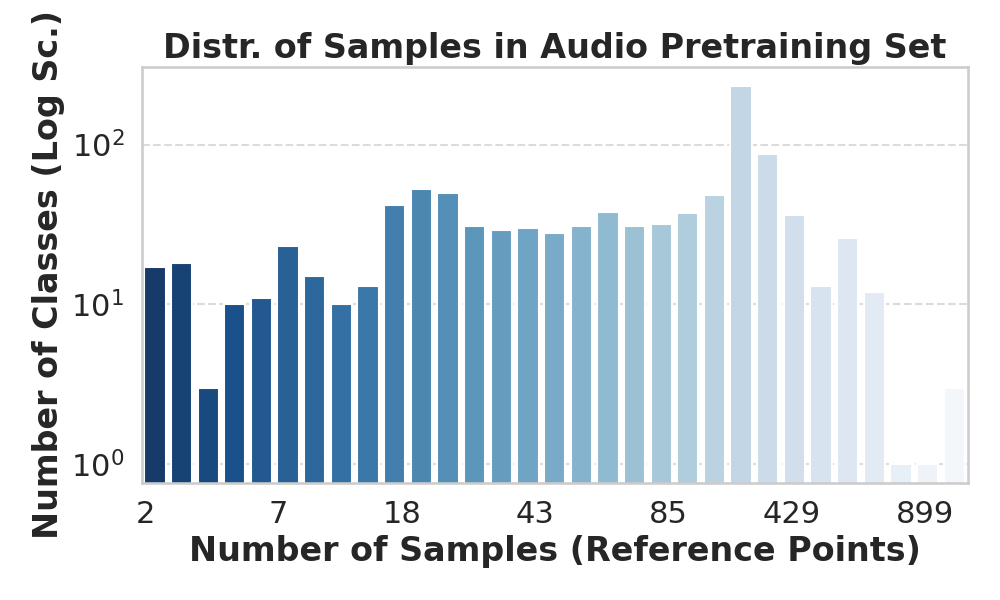}
        
         \caption*{ \textbf{(b)} Distribution of audio samples per class in the pretraining set, spanning >50K audio recordings. 
        }
    \end{subfigure}
    
    
    \begin{subfigure}[b]{0.24\textwidth}
        \centering
        \includegraphics[width=\textwidth]{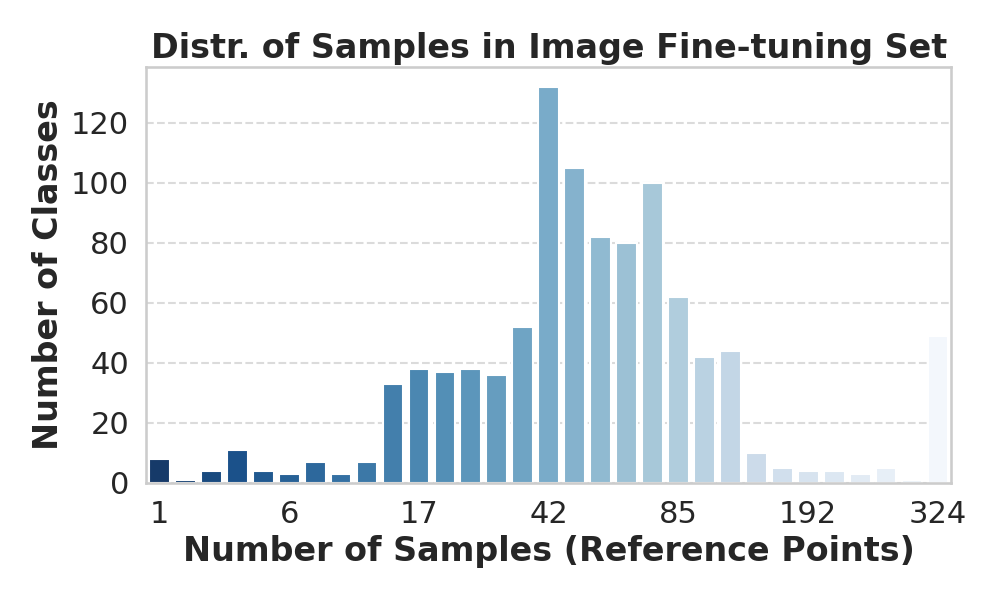}
      \caption*{ \textbf{(c)} Number of image samples per class in the fine-tuning set, consisting of 83K samples. 
        }
    \end{subfigure}
    \hfill
    \begin{subfigure}[b]{0.23\textwidth}
        \centering
        \includegraphics[width=\textwidth]{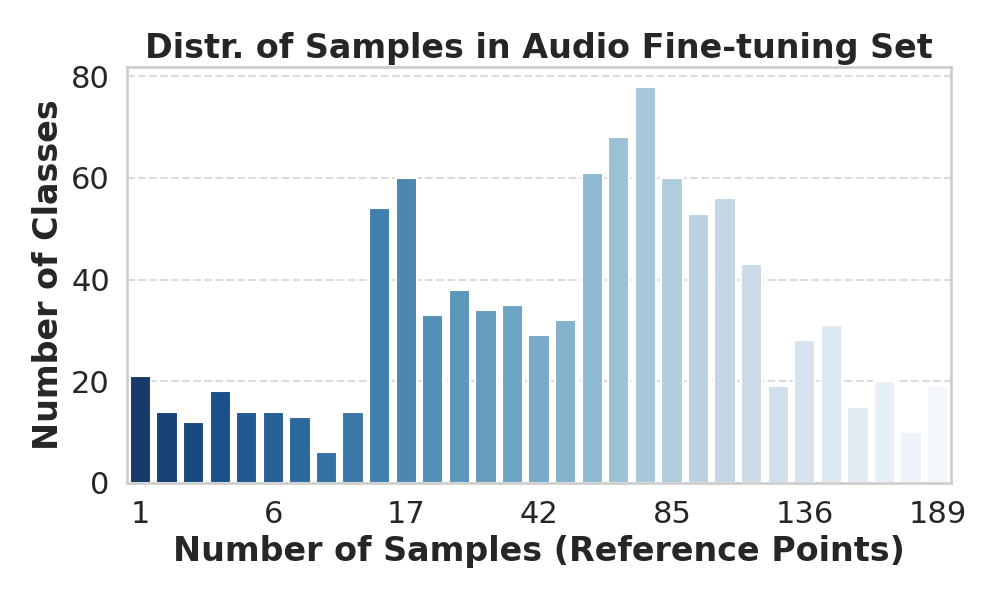}
      \caption*{ \textbf{(d)} Number of audio samples per class in the fine-tuning set with a total of 65K samples.}
        
    \end{subfigure}
    \vspace{-1ex}
    \caption{\textbf{Distribution of Image and Audio Samples} in the Pretraining and Fine-Tuning Sets.}
    \label{fig:statistics}
\end{figure}

Key statistics of the MAviS-Dataset highlight its scale and comprehensiveness. The dataset contains approximately \textbf{420,000 images} and \textbf{115,000 audio clips}, representing \textbf{1,013 bird species} worldwide and covering all major avian families and geographical regions.
Detailed class-wise distributions are illustrated in Figure~\ref{fig:statistics}. 
On average, each species is associated with roughly 210 images and 115 audio recordings, although the number varies by species. Some of the species have between 5-10 images and 1-5 recordings, while some highly documented birds have significantly more. The audio collection alone amounts to thousands of hours of bird vocalizations. 
Every species is supplemented with a carefully vetted textual description, typically consisting of multiple paragraphs detailing its visual and sound characteristics, habitat, and behavior. All species samples include the three modalities, strengthening MAviS as a robust multimodal dataset.

MAviS-Dataset surpasses prior fine-grained benchmarks in data volume. For instance, the iNaturalist 2021 competition, a benchmark dataset for species classification, included $\sim$1,500 bird species in its vision task~\cite{NEURIPS2024_ef3713b8}. In contrast, MAviS, despite covering 1,013 species, contains a significantly larger number of samples per species, providing a more balanced and comprehensive dataset for fine-grained bird classification. 




\section{MAviS-Bench: An Evaluation Benchmark for Avian Understanding}
\label{sec:bench}

To systematically evaluate model performance, we propose MAviS-Bench, a benchmark with more than 3,900 curated evaluation samples resulting in over 25,000 instruction-response pairs. 


\subsection{Benchmark Criteria}

MAviS-Bench is designed to evaluate multimodal models on fine-grained species understanding tasks by assessing both perception-based classification and reasoning-driven inference. Inspired by MMBench~\cite{liu2024mmbenchmultimodalmodelallaround}, the benchmark includes two primary task categories:

\begin{itemize}
    \item 
    \textbf{Perception Tasks:} 
    These tasks evaluate the model’s ability to classify bird species based on visual or auditory inputs and perform multimodal retrieval. The goal is to assess how well a model can distinguish fine-grained species differences.
    
    \item \textbf{Reasoning Tasks:} 
    Beyond classification, reasoning tasks require a contextual understanding of multimodal data. These include multimodal question answering and caption generation, as well as testing whether models can infer knowledge rather than rely solely on pattern recognition.
\end{itemize}



Each sample in the benchmark includes at least one targeted question testing species identification, reinforcing the hypothesis that generalized LLMs struggle with fine-grained classification. To push models beyond explicit species labeling, 30\% of the generated instructions deliberately omit the bird’s name, requiring the model to infer species-related attributes such as habitat, behavior, and ecological roles purely from contextual cues. The 30\% omission ratio was selected both to create a balanced dataset with varying levels of difficulty and to enable controlled ablation studies that test the model’s capacity to generalize beyond species names. We define these as \textit{hard questions}, since they demand greater reasoning effort over fine-grained visual and acoustic traits, 
without relying on surface-level lexical shortcuts. Omitting the species name encourages grounding-based reasoning and ensures that the model cannot answer using memorized associations. For instance, a query may ask for the specific call type in a given audio input, which cannot be resolved with text-only cues, even if the species name is known. The difficulty increases significantly when the model must infer such traits without access to the species label, thereby requiring deeper cross-modal understanding and reasoning.



\subsection{Benchmark Composition}

To maintain openness and reproducibility, MAviS-Bench was curated exclusively from publicly available sources -- BirdCLEF, Rare-birds, and Tree-of-Life-10M. This allows MAviS-Bench to be shared without licensing restrictions, ensuring broad accessibility. 
Almost every class in the dataset is represented by two samples per modality, which comprises 1,892 audio samples, 2,017 image samples, and approximately 25,000 question--answer pairs. This ensures a balanced multimodal distribution, facilitating a detailed assessment of the model's performance in images, audio, and text queries, thus providing a controlled and scalable benchmarking framework for multimodal species understanding.

\section{Benchmarking and Experimental Evaluation}

\subsection{Baseline Model Selection}

The evaluation of multimodal models requires a robust benchmarking and fine-tuning process to systematically assess their ability to integrate and process heterogeneous data modalities. 
To ensure a comprehensive and balanced evaluation, we consider state-of-the-art multimodal models, including GPT-4o-mini~\cite{openai2024gpt4ocard} and Gemini 1.5~\cite{geminiteam2024gemini15unlockingmultimodal} (proprietary),
as well as Phi-4~\cite{abdin2024phi4technicalreport} 
and MiniCPM-o-2.6~\cite{tr-2.6} (open source). 


To investigate the benefits of domain-specific adaptation, we introduce {MAviS-Chat}, a baseline initialized from MiniCPM-o-2.6.
It integrates the following components: \texttt{SigLip-400M} (vision encoder), \texttt{Whisper-medium-300M} (audio encoder), and \texttt{Qwen2.5-7B} (language model), for a total parameter count of $\sim$8B. 
MAviS-Chat is fine-tuned on the MAviS-Dataset using both pretraining and instruction subsets, incorporating image/audio, and text modalities. For adaptation, we adopt the LoRA framework~\cite{hu2022lora} following the configuration of MiniCPM-o-2.6.
We provide more discussions about the layer selection and training schedules in \S~\ref{sec:ablation}.


\subsection{Evaluation Metrics}

The evaluation of MAviS-Chat and baseline models is carried out using lexical overlap metrics (BLEU, ROUGE-1, METEOR) and embedding-based semantic similarity metrics (BERTScore~\cite{zhang2019bertscore}, MoverScore~\cite{zhao2019moverscore}).
Lexical overlap metrics focus on the degree of direct word or n-gram matches between generated and reference texts. They provide a straightforward measure of surface-level similarity, which is particularly useful in tasks where exact wording is critical, while Embedding-based semantic similarity metrics utilize vector representations of text to capture deeper semantic relationships. By leveraging contextual embeddings, they assess the meaning conveyed by the text, offering a more nuanced evaluation of language generation models.

To operationalize MAviS-Bench, we further introduce MAviS-Eval (`ME'), a reference-based scoring framework inspired by the RecEval~\cite{prasad2023recevalevaluatingreasoningchains} metric suite.
Unlike reference-free or embedding-based metrics, MAviS-Eval uses a structured rubric tailored to the ecological context and perceptual nature of current tasks. Each model's output is assessed using a seven-dimensional scoring rubric that includes correctness, modality grounding, conciseness, clarity, confidence appropriateness, hallucination, and coverage.
Each attribute is rated on a 1-5 scale. The average of these dimensions yields an ``Overall ME Score'', allowing robust cross-model comparisons. Importantly, MAviS-Eval addresses one of the key weaknesses of LLM-as-a-judge by controlling for verbosity bias and hallucination tolerance, ensuring more reliable and domain-aligned evaluations. 
The detailed structure of MAviS-Eval can be found in Appendix~\ref{sec:appendixB}.

\begin{table}[t]
    \centering
    \renewcommand{\arraystretch}{1.3}
    \setlength{\tabcolsep}{4pt}
    \large
    \resizebox{\linewidth}{!}{
    \begin{tabular}{lcccc|cc}
        \hline
        \rowcolor{myblue}
        
        \textbf{Model} & \textbf{R1} & \textbf{M} & \textbf{BS} & \textbf{MS} & \textbf{ME-a}  & \textbf{ME-c} \\
        \hline
        GPT-4o & 30.55 & \textbf{34.08} & \underline{87.92} & 54.03 & 59.88  & \textbf{71.18} \\
        GPT-4o-mini & 24.28 & 29.72 & 86.52 & 52.58 & \underline{60.04}  & \underline{69.16} \\
        Gemini 1.5 & 18.95 & 23.31 & 84.42 & 50.63 & 45.16 & 47.36 \\
        \midrule
        Phi-4-MM-Instruct & \underline{32.43} & \underline{30.18} & \textbf{88.68} & \underline{54.42} & 48.04 & 57.70\\
        MiniCPM-o-2.6 & 19.77 & 27.14 & 85.33 & 52.10 & 55.46 & 52.42 \\
        \midrule
        \textbf{MAviS-Chat (our)} & \textbf{34.17} & 29.31 & 87.42 & \textbf{54.76} & \textbf{61.10} & 59.92 \\
        \hline
    \end{tabular}}
    \vspace{-1ex}
    
    \caption{\textbf{Performance comparison on MAviS-Bench.} 
    `R1', `M', `BS', and `MS' denote ROUGE-1, METEOR, BERTScore, and MoverScore. 
    `ME-a' denotes MAviS-Eval (audio), generated from model responses to audio only questions.
    `ME-c' is the MAviS-Eval (combined), 
    evaluated on the full set comprising both audio- and image-based test samples.
    }
    \vspace{-1ex}
    \label{tab:benchmark_results}
\end{table}

\subsection{Evaluation Results on MAviS-Bench}\label{sec:sota_evaluation}

Table~\ref{tab:benchmark_results} reports the performance of leading multimodal models on MAviS-Bench, evaluated using conventional generation metrics and the proposed MAviS-Eval metric. All baseline models were evaluated in zero-shot or few-shot settings. This setup ensures a fair performance comparison and highlights the effects of domain-specific instruction tuning used for MAviS-Chat.

\noindent\textbf{GPT-4o} achieves the highest MAviS-Eval (`ME-a/c') scores and strong semantic alignment, but exhibits several drawbacks. We found that around 5\% of its outputs lack informative content, while others rely on vague descriptors without integrating modality-specific cues. Its tendency to default to common species reflects a bias from pretraining, limiting its precision in rare or ambiguous cases.

\noindent\textbf{GPT-4o-mini} and \textbf{Gemini 1.5} perform similarly across semantic metrics but show weaker results in MAviS-Eval and lexical overlap, indicating reduced capacity for detailed, fine-grained reasoning.

\noindent\textbf{Phi-4-MM-Instruct} ranks highest across standard lexical metrics and BERTScore (88.68), but struggles with unimodal audio inputs—frequently producing uncertain or evasive responses. 
This limits its effectiveness in grounded audio understanding.

\noindent\textbf{MiniCPM-o-2.6} offers a well-balanced profile, with strong semantic scores and reliable performance in audio-based reasoning. However, with approximately 15B parameters and limited open-access APIs, it remains relatively expensive to use at scale. Its lightweight architecture still makes it a feasible candidate for field deployments when resources permit.

\noindent\textbf{MAviS-Chat}, our three-step fine-tuned model, achieves high scores in ROUGE-1 and MoverScore, while also surpassing open-source models, Phi-4-MM-Instruct and MiniCPM-o-2.6, on MAviS-Eval metric. These results reflect improved grounding and fine-grained multimodal reasoning. 

In summary, while general-purpose models like GPT-4o (with over 1T parameters) demonstrate strong capabilities, their large scale and high inference cost limit their practicality for real-world ecological applications. MAviS-Chat offers a promising middle ground - delivering targeted performance gains through domain-specific fine-tuning while maintaining flexibility for low-cost deployment in conservation scenarios.

\begin{table}[t]
    \centering
    \renewcommand{\arraystretch}{1.3}
    \setlength{\tabcolsep}{5pt}
    \footnotesize
    \resizebox{\linewidth}{!}{
    \begin{tabular}{lcccccc}
    \toprule
    \rowcolor{myblue}
    
    \textbf{Model} & \textbf{B1} & \textbf{R1} & \textbf{M} & \textbf{BS} & \textbf{MS} & \textbf{ME} \\
    \midrule
    GPT-4o-mini & 10.27 & 11.53 & 14.49 & 83.24 & 51.06 & 31.06 \\
    \textbf{MAviS-Chat}  & \textbf{18.55} & \textbf{19.42} & \textbf{19.04} & \textbf{85.02} & \textbf{51.18} & \textbf{39.30} \\
    \bottomrule
    \end{tabular}}
    \vspace{-2ex}
    
    \caption{\textbf{Results on the real-world audio-visual video evaluation set.} 
    `B1', `R1', `M', `BS', and `MS' denote BLEU-1, ROUGE-1, METEOR, BERTScore, and MoverScore. 
    `ME' denotes MAviS-Eval.}
    \vspace{-1ex}
    
    \label{tab:video-benchmark}
\end{table}

\subsection{Evaluation on Real-World Videos} 
To further quantify multimodal reasoning, we also construct a real-world audio-visual evaluation set, 
comprising wildlife video recordings shared by the Cornell Lab of Ornithology (the video assets will not be released given licensing requirements, but we will provide the video ids for reference). These videos are not synthetic composites; they contain naturally synchronized auditory and visual signals. 
Each bird species class in this benchmark is represented by at least two distinct videos. Each video is paired with seven referential questions: two visual-based, two auditory-based, and three temporal/event-based. This evaluation set allows us to assess how well a model integrates information across time and modalities.
We compare our method with GPT-4o-mini, a cost-efficient yet competitive baseline.
As shown in Table~\ref{tab:video-benchmark}, our MAviS-chat outperforms GPT-4o-mini across all metrics.
These results further highlight the effectiveness and superiority of our model in real-world multimodal avian analysis.

\subsection{Ablation Studies on Training Strategies}\label{sec:ablation}

\noindent\textbf{Impact of Different LoRA Tuning Layers.}
In the evaluation of MAviS-Chat, various architectural decisions, including the inclusion of LoRA layers, were found to have significant effects on the model's multimodal reasoning ability and text generation quality. 
Table~\ref{tab:tuning_lora_layers} reveals how the choice of architectural layers impacts both model alignment with multimodal data and reasoning capability, as assessed by MAviS-Eval.

The \textit{All Projectors (our model)} variant, in which all the linear projectors are trainable,
significantly improves the MiniCPM-o-2.6 baseline, e.g., achieving a high MAviS-Eval score of 59.92. 
This configuration, which integrates projection layers into the architecture, leverages their ability to capture and align semantic features across modalities, likely contributing to its superior performance in reasoning. 
The LoRA layers, embedded within this architecture, facilitate effective alignment and enhance the model’s capacity to interpret multimodal input with strong semantic grounding. 

\begin{table}[t]
    \centering
    \renewcommand{\arraystretch}{1.3}
    \setlength{\tabcolsep}{5pt}
    \footnotesize
    \resizebox{\linewidth}{!}{
    \begin{tabular}{lcccccc}
        \toprule
        \rowcolor{myblue}
        
        \textbf{Training Layers} & \textbf{B1} & \textbf{R1} & \textbf{M} & \textbf{BS} & \textbf{MS} & \textbf{ME} \\
        \midrule
        All Proj. (MiniCPM.) & 11.43 & 19.77 & 27.14 & 85.33 & 52.10 & 52.42 \\
        \midrule
        \textbf{All Proj. (our model)} & 25.98 & 34.17 & \textbf{34.85} & \textbf{87.42} & 54.76 & \textbf{59.92} \\
        FFN only (our model) & \textbf{39.65} & \textbf{36.33} & 29.61 & 86.72 & \textbf{55.11} & 59.42 \\
        \bottomrule
    \end{tabular}}
    \vspace{-2ex}
    
    \caption{\textbf{Ablation on the layers tuning with LoRA.} 
    `B1', `R1', `M', `BS', `MS', and `ME' denote BLEU-1, ROUGE-1, METEOR, BERTScore, MoverScore, and MAviS-Eval (combined). 
    The first two rows denote that all projectors are fine-tuned using LoRA adapters, whereas the last row corresponds to training only the Feed-Forward Network (`FFN').}
    \vspace{-1ex}
    
    \label{tab:tuning_lora_layers}
\end{table}

In contrast, the \textit{Feed-Forward Network (FFN) only} variant, which exclusively integrates LoRA layers to the FFN, 
shows strong performance in text generation metrics like ROUGE-1 and BLEU-1. However, it still falls behind on the MAviS-Eval compared to the all projector tuning strategy, 
demonstrating that while LoRA layers can improve text generation, they may not necessarily enhance cross-modal reasoning unless coupled with a robust fusion strategy like projection layers. 
These findings underscore the critical role of tuning layers in determining how effectively the model fuses multimodal information for ecological reasoning.

\begin{table}[t]
    \centering
    \renewcommand{\arraystretch}{1.2} 
    \setlength{\tabcolsep}{3pt} 
    
    \footnotesize 
    \resizebox{0.48\textwidth}{!}{

    \begin{tabular}{
        p{0.2\linewidth} 
        p{0.15\linewidth} 
        c c c c c c
    }
        \toprule
        \rowcolor{myblue}
        
        \textbf{Training Strategy} & \textbf{Data Setup} & \textbf{B1} & \textbf{R1} & \textbf{M} & \textbf{BS} & \textbf{MS} & \textbf{ME} \\
        \midrule
        MiniCPM & - & 11.43 & 19.77 & 27.14 & 85.33 & 52.10 & 52.42 \\
        \midrule
        
        \textbf{Three-step Tuning} & \textbf{A}\textrightarrow \textbf{I}\textrightarrow\textbf{A} (full) & 25.98 & 34.17 & \textbf{34.85} & \textbf{87.42} & 54.76 & \textbf{59.92} \\
        {Two-step Tuning} & A\textrightarrow I (full) & \textbf{38.88} & \textbf{36.46} & 29.31 & 86.66 & \textbf{55.07} & 58.16 \\ \midrule
        Small+VQ Proj. & A\textrightarrow I\textrightarrow A (small) & 30.64 & 33.99 & 34.07 & 87.35 & 54.58 & 58.84 \\
        \bottomrule
    \end{tabular}
    }
    \vspace{-2ex}
    \caption{\textbf{Ablation on training data setups.} `B1', `R1', `M', `BS', `MS', and `ME' denote BLEU-1, ROUGE-1, METEOR, BERTScore, MoverScore, and MAviS-Eval (combined), respectively. The last three rows denote our MAviS-Chat model. \textbf{A/I}: Audio/Image. \textbf{full/small}: 65K (audio)/83K (image) or 35K data.}
    \vspace{-1ex}
    
    \label{tab:mavis_variations}
\end{table}

\noindent\textbf{Training Data Order in Fine-tuning.}
The training sequence also plays a crucial role in improving MAviS-Chat’s performance. 
Our default model, {MAviS-Chat}, was trained using a \textit{three-step} sequential fine-tuning strategy: 1) audio-based fine-tuning on 65k Q\&A pairs, 2) image-based fine-tuning on 83k Q\&A pairs, and 3) a final audio-based fine-tuning pass to restore acoustic grounding after the visual stage. 
A comparative analysis of training strategies shows that sequential data presentation impacts both lexical quality and multimodal reasoning ability (Table~\ref{tab:mavis_variations}).

The \textit{Three-Step Fine-tuning} variant stands as the most effective strategy, achieving the highest performance overall, including both standard generation metrics and the MAviS-Eval score. This method begins with audio inputs, followed by image, and concludes with a final realignment to audio data. This progressive fine-tuning approach promotes structured adaptation across modalities, enabling the model to better align visual and auditory signals. The sequential exposure to image and audio data enhances the model's ability to process multimodal signals in a more coherent and grounded manner, as opposed to models that do not realign at the final step. This method demonstrates that the order and sequencing of data can significantly impact the model’s overall performance in multimodal tasks.

In comparison, the \textit{Two-Step Fine-tuning} strategy, where the model is not realigned to audio data in the final step, still achieves strong performance (MAviS-Eval 58.16) but falls short of the Three-Step variant. The additional step in the Three-Step fine-tuning leads to a noticeable boost in both lexical quality (e.g., METEOR 34.85 vs. 29.31) and multimodal alignment, further emphasizing that training data sequencing is critical for achieving high performance in multimodal reasoning tasks, additional results can be found in appendix \ref{sec:appendixC}.

\noindent\textbf{Training Data Size.} The impact of dataset size on model performance is also notable. As shown in Table~\ref{tab:mavis_variations}, despite being trained on a smaller dataset (35K samples), the \textit{Small Sample + VQ projectors} variant achieved a competitive MAviS-Eval score of 58.84, showcasing that a smaller, more domain-specific dataset can still effectively improve multimodal alignment when combined with task-specific fine-tuning strategies. This result highlights the importance of quality over quantity in data, suggesting that a well-curated, smaller dataset focusing on relevant task-specific examples can outperform a larger dataset without such focus.



\vspace{-0.5ex}
\section*{Conclusion}

To translate AI's promise into tangible conservation tools, we introduce the MAviS suite. Our work provides a complete, domain-specific ecosystem, uniting a dataset, model, and benchmark, to build the essential infrastructure for automated, large-scale avian understanding. This research lays the groundwork for deploying practical AI systems in the wild, offering a scalable blueprint for developing similar specialized tools to safeguard diverse species within the world's vulnerable ecosystems.



\vspace{-0.5ex}
\section*{Limitations}
While the evaluated models demonstrate strong performance on multimodal avian understanding, several challenges remain. Rare species with limited data are still difficult to recognize reliably, and audio-based reasoning tasks are prone to modality-specific biases and environmental noise. Additionally, model performance is constrained by the scale and diversity of current training data, as well as available computational resources. Future work should explore strategies for improving generalization to unseen species, enhancing robustness under varied acoustic or visual conditions, expanding coverage of long-tailed categories, and incorporating video data for fine-tuning. Fine-tuning on video, which integrates both visual and audio modalities, may significantly impact model performance by providing richer temporal and spatial context, leading to better handling of complex scenarios and improved recognition of dynamic behaviors.


\section*{Acknowledgments}
We would like to thank the contributors and maintainers of The Macaulay Library at the Cornell Lab of Ornithology, BirdCLEF, Tree-of-Life, and iNaturalist for providing the invaluable data used in this study.
This work is partially supported by the \textit{Google Research Award}, \textit{Meta Regional Research Grant 2025}, and the \textit{NVIDIA Academic Grant 2025}.
We also declare no competing interests related to this work.

\bibliography{custom}

\appendix

\newpage
\section{Prompt used in Fine-tuning set generation}\label{sec:appendixA}
\begin{figure*}[tp]
    \centering
    \includegraphics[width=\textwidth]{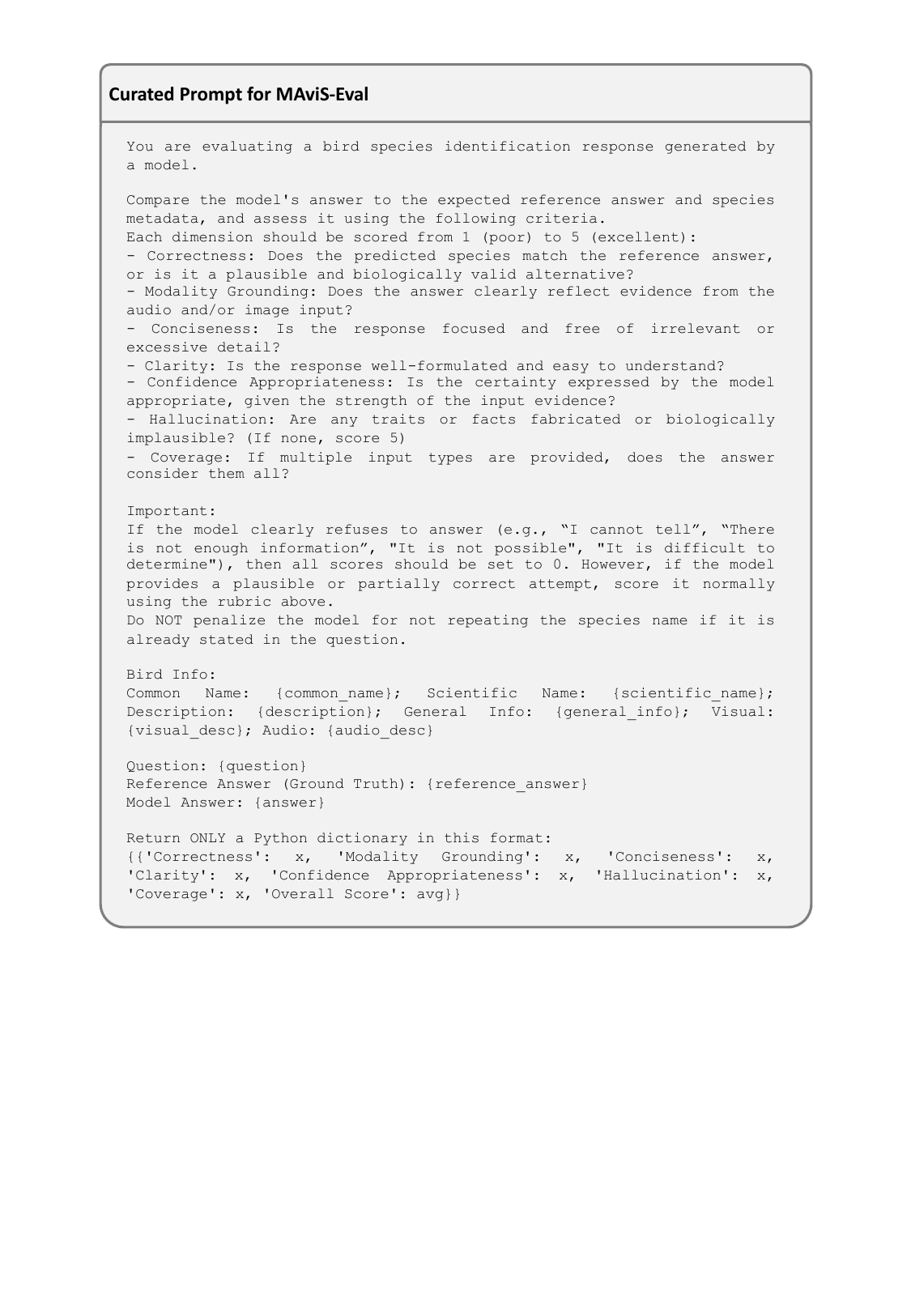} 
    
    \caption{Prompt used for the instruction set generation for audio and video inputs.}
    \label{fig:eval_prompt}
\end{figure*}

Figure~\ref{fig:eval_prompt} displays the detailed prompt used to generate question-answer pairs for the fine-tuning set.

\section{MAviS-Eval Prompt}
\label{sec:appendixB}
\begin{figure*}[tp]
    \centering
    \includegraphics[width=\textwidth]{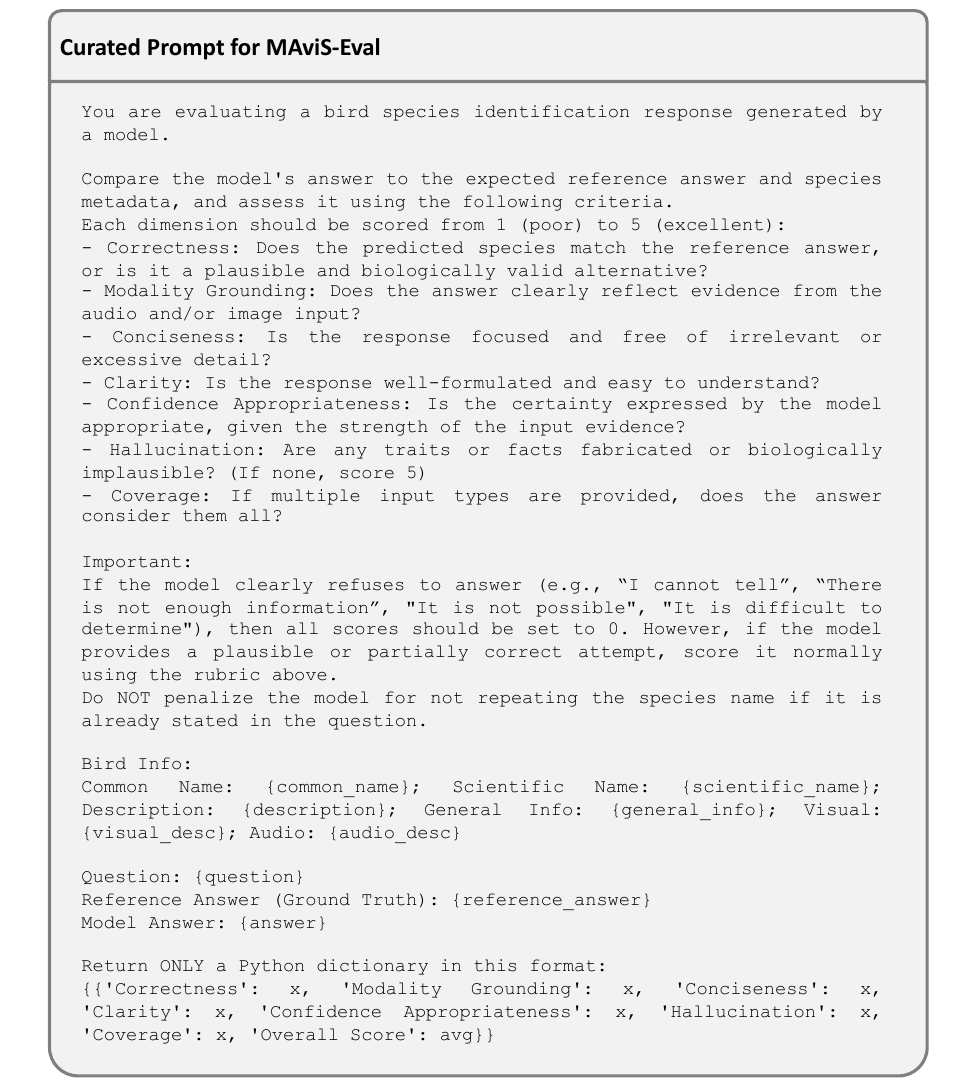} 
    \caption{Prompt template used in the MAviS-Eval metric to evaluate model performance on avian species identification within MAviS-Bench.}
    \label{fig:mavis-eval}
\end{figure*}

The MAViS-Eval framework is designed to evaluate the reasoning quality of multimodal models across text, image, and audio inputs. It employs a structured system prompt, as shown in Figure~\ref{fig:mavis-eval}, that guides the assessment of model outputs based on multiple dimensions, including faithfulness, informativeness, repetition, hallucination, redundancy, semantic coverage, reasoning alignment, commonsense, and completeness of reasoning steps. Each aspect is rated on a scale from 1 to 5, with detailed evaluation criteria provided to ensure consistency and objectivity across annotators. This framework enables a fine-grained analysis of a model’s ability to generate coherent, accurate, and semantically rich responses in complex, multi-modal scenarios.

\section{Lexical Quality Improvement Evidence}
\label{sec:appendixC}
Table~\ref{tab:lexical-quality} presents more detailed lexical quality evaluation results for our Two-step and Three-step fine-tuning methods, measured using ROUGE and METEOR scores. These results provide evidence of the improvements in lexical quality achieved by our approach.

\begin{table}[h]
   \setlength{\tabcolsep}{5pt}
    \renewcommand{\arraystretch}{1.5}
    \centering
    \Huge
    \resizebox{0.48\textwidth}{!}{
\begin{tabular}{lccccc}
\toprule
\textbf{Methods} & \textbf{ROUGE-1} & \textbf{ROUGE-2} & \textbf{ROUGE-L} & \textbf{ROUGE-W} & \textbf{METEOR} \\
\midrule
Two-step   & \textbf{36.46} & 36.21 & 15.13 & 29.56 & 29.31 \\
Three-step & 34.17 & \textbf{37.33} & \textbf{16.40} & \textbf{31.36} & \textbf{34.85} \\
\bottomrule
\end{tabular}}
\caption{\textbf{Lexical quality evaluation} of Two-step vs. Three-step tuning methods across ROUGE and METEOR metrics. Bold values indicate the best scores.}
\label{tab:lexical-quality}
\end{table}

\section{Additional Examples of Questions}
\label{sec:appendixD}

\begin{figure}[h]
    \centering
    \includegraphics[width=1.0\linewidth]{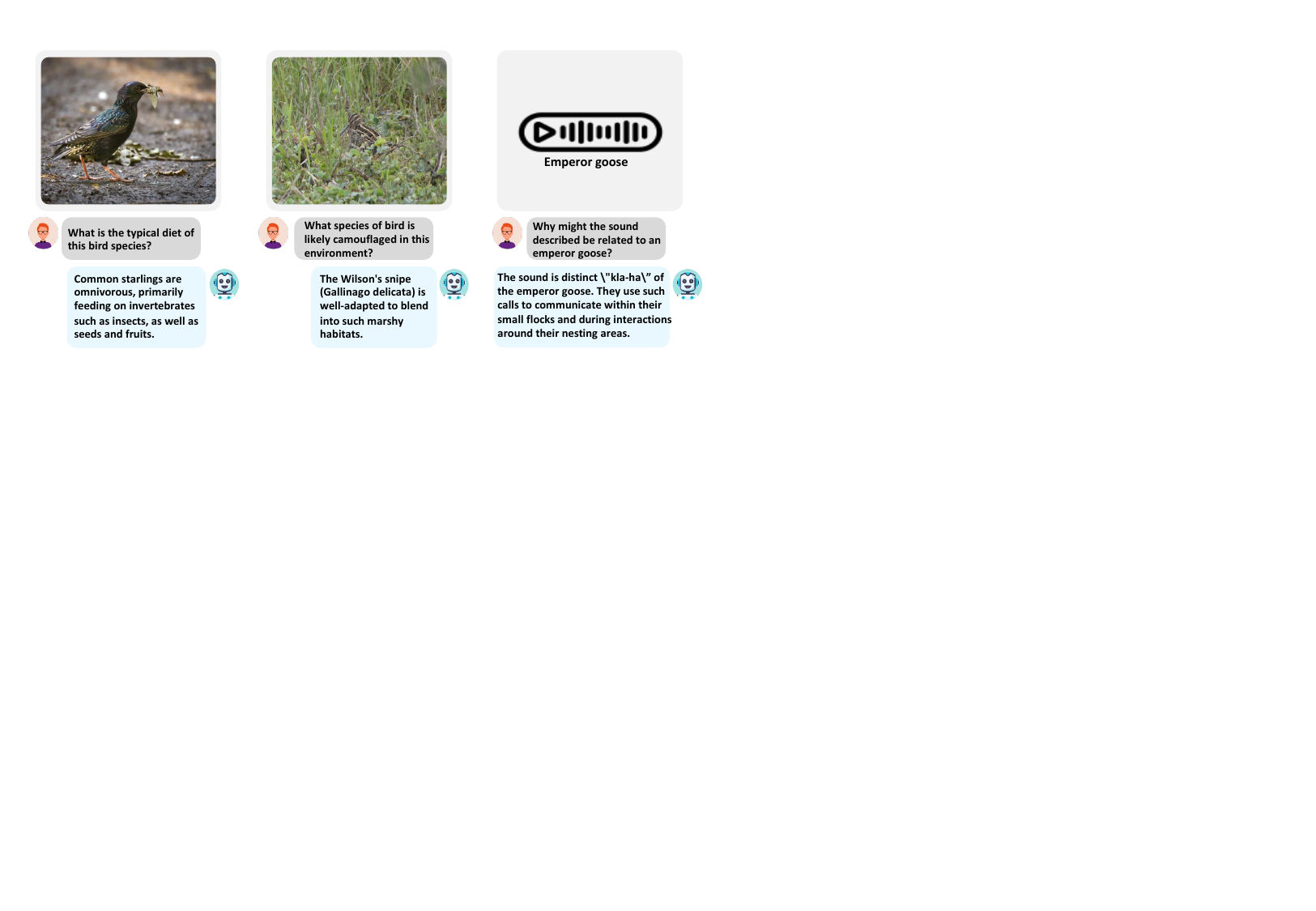}
    \caption{Additional multimodal question–answer samples from MAviS-Dataset, covering ecological inference such as eating habits, visual behaviour, and appearance interpretation, as well as detailed understanding of bird emotional tone and calling types from audio.
    }
    \label{fig:appendix_samples}
\end{figure}

In this appendix, we provide additional qualitative examples of question–answer pairs from MAviS-Dataset. 
Figure~\ref{fig:appendix_samples} illustrates two image-based samples and one sound-based sample. 
These examples highlight the diversity of modalities and the reasoning depth required to answer proposed questions.

\section{List of audio recordings received from the Macaulay Library}
\label{sec:appendixE}

In accordance with the Macaulay Library’s data usage policy, we provide a detailed list of the recordings used from the Cornell Lab of Ornithology as follows:
ML109293, ML136339, ML136341, ML136382, ML136384, ML136437, ML137535, ML137538, ML137575, ML137577, ML163310, ML207751, ML239100, ML32677601, ML32677611, ML32677621, ML32677631, ML32677751, ML32677761, 
ML32677771, ML32677781, ML32677791, ML32677891, ML79092981, ML167006351, ML167947021, ML167947041, ML187278491, ML199008281, ML203903841, ML203903851, ML286841351, ML288189241, ML346116351, ML419406581, ML461652931, ML464492891, ML464492901, ML464492911, ML475249271, ML516015061, ML516188351, ML574964731, ML580949301, ML607945611, ML616201979, ML618419611, ML618419630, ML620697520, ML621291007, ML623836828, ML624691011, ML628016447, ML628016448, ML628016449, ML2064, ML128577511, ML203887571, ML282851631, ML285177021, ML369376051, ML383938811, ML383938821, ML385997171, ML484545911, ML484546021, ML484546081, ML485439021, ML487658371, ML508652871, ML508798331, ML611601110, ML624419602, ML625389508, ML625389509, ML625389510, ML626862622, ML632433568, ML6552, ML6554, ML6555, ML6556, ML27220, ML27227, ML37779, ML55183, ML55218, ML57351, ML73477, ML73479, ML74032, ML76679, ML81869, ML81870, ML185778, ML214495, ML214582, ML225516, ML225521, ML249384, ML299978, ML299990, ML299991, ML300124, ML83978491, ML89672471, ML90673081, ML91484641, ML140333091, ML143939971, ML145613381, ML147179141, ML150284861, ML155747601, ML155747771, ML165385601, ML166887881, ML172847601, ML203939381, ML211416411, ML218331611, ML228006651, ML244852691, ML292346411, ML300684111, ML312684421, ML313036481, ML336195631, ML357417461, ML358255441, ML394829311, ML394829321, ML419482601, ML421560791, ML422055701, ML422732731, ML435932161, ML437005601, ML479543831, ML508509161, ML512208261, ML512208271, ML524864671, ML531510561, ML541637891, ML561775651, ML561775661, ML569976201, ML579429631, ML579614361, ML592743711, ML606269691, ML609056384, ML610758317, ML614640712, ML614640713, ML614646683, ML614646906, ML614813346, ML615568830, ML616093362, ML616103766, ML616131913, ML616264676, ML616774354, ML616776290, ML625817139, ML628927965, ML629478601, ML629478602, ML629478603, ML629478604, ML629478605, ML632206745, ML632208026, ML632268760, ML632327499, ML632810562, ML613, ML70614, ML70615, ML138332, ML200516, ML200575, ML20736241, ML100661331, ML104892821, ML113371101, ML130628961, ML135623341, ML135623401, ML203923781, ML203967001, ML326759541, ML360483431, ML360701741, ML364809191, ML383989931, ML428972741, ML439852461, ML442430261, ML460680091, ML466211131, ML470665901, ML483741541, ML507284941, ML534711261, ML566723681, ML609070366, ML609376015, ML620753843, ML620944639, ML621139521, ML621139582, ML622226425, ML627682015, ML627682016, ML627682017, ML631237163, ML632694386, ML140383, ML140384, ML140385, ML90121461, ML170760231, ML614644638, ML614793411, ML614793412, ML615442529, ML615442544, ML627532241, ML631019171, ML631862720, ML631862721, ML631862722, ML631862723, ML4213, ML67341, ML67342, ML67344, ML67346, ML67347, ML77097, ML77099, ML77200, ML77209, ML77210, ML77212, ML77213, ML79426, ML79427, ML79428, ML79429, ML79430, ML79431, ML79432, ML100908, ML136157, ML140079, ML195711, ML197068, ML20082611, ML29804851, ML30325331, ML41103151, ML49767391, ML61356231, ML84962911, ML95384151, ML96418831, ML100484441, ML106879581, ML106887201, ML159384641, ML159407151, ML162243691, ML163165391, ML163589661, ML164263741, ML164318071, ML164488731, ML164488831, ML164948831, ML165501321, ML167006181, ML167157871, ML167696541, ML167696571, ML204009911, ML243717431, ML243717591, ML243717601, ML243830421, ML244525301, ML245382991, ML247192231, ML247192411, ML250014471, ML253115981, ML337423361, ML349138061, ML349559071, ML350043131, ML350379691, ML351535231, ML352682711, ML353314341, ML353656551, ML359603191, ML406824651, ML455284911, ML455313531, ML459038791, ML461943481, ML473093401, ML575240291, ML586947261, ML587797761, ML592484661, ML594307791, ML597664361, ML599419441, ML601942601, ML608380147, ML608902921, ML619123632, ML619123633, ML620351022, ML620382349, ML620383141, ML620383624, ML620599348, ML621295005, ML621647591, ML624215786, ML629152493, ML38329, ML38334, ML40560, ML51052, ML105222, ML105223, ML105230, ML105236, ML105505, ML105517, ML105538, ML110146, ML110148, ML110151, ML110152, ML110153, ML110154, ML110155, ML141114, ML141122, ML147512, ML147587, ML188098, ML188204, ML188207, ML28259771, ML28843051, ML32304681, ML55399831, ML62609671, ML62682521, ML65177491, ML66076691, ML66076741, ML100649261, ML101142431, ML108818221, ML147506301, ML149231221, ML151148141, ML151148371, ML159367311, ML161342921, ML166964811, ML166964881, ML171598251, ML225152221, ML228695511, ML240072311, ML246217611, ML246217731, ML260751891, ML297631471, ML297670401, ML297688491, ML298001821, ML302060251, ML302307531, ML326058611, ML326719731, ML326842251, ML331286831, ML331758981, ML332706641, ML332875191, ML334291891, ML334944891, ML334945241, ML336039641, ML336921471, ML341788561, ML345427061, ML349113261, ML349516481, ML355799461, ML355799561, ML408882971, ML408886521, ML408982791, ML430107911, ML431689101, ML438777691, ML442053701, ML447790341, ML535722541, ML564017201, ML565431621, ML565436301, ML573933871, ML573937151, ML574667561, ML575624611, ML601723231, ML613639565, ML613639568, ML613639569, ML613639570, ML618672213, ML618672300, ML619195119, ML2960, ML143565231, ML203938181, ML204012501, ML301926801, ML356483451, ML534286231, ML613907633, ML614060433, ML614159611, ML614164395, ML614165268, ML614314788, ML614544552, ML620526514, ML622060366, ML622956127, ML624580306, ML624580353, ML627481357, ML628647618, ML628647619, ML629926483, ML631073140, ML631133517, ML631511891, ML631511892, ML631511893, ML631511895, ML632452192, ML71386, ML71389, ML71392, ML162909, ML110009741, ML184446301, ML184448291, ML185204731, ML185204821, ML191037341, ML191850361, ML203916091, ML269675871, ML278594371, ML383052321, ML464123631, ML477317761, ML489539701, ML489539711, ML495316361, ML512968171, ML519869651, ML525153411, ML533837831, ML535017951, ML535018031, ML535018061, ML535018091, ML536389431, ML558085041, ML560557461, ML572371481, ML609067244, ML610891254, ML612132770, ML617870651, ML617870652, ML618709640, ML619476184, ML620202006, ML620202094, ML621024619, ML621024633, ML621373697, ML624879235, ML625718303, ML625821511, ML627779105, ML629269390, ML631606862, ML3397, ML365642801, ML367300481, ML367301691, ML367301711, ML375405451, ML375405761, ML375504261, ML375504641, ML375508541, ML375509601, ML375509891, ML375510081, ML375510231, ML378084131, ML378085941, ML378087181, ML378088941, ML381363681, ML381363741, ML383029571, ML383029731, ML383029881, ML383051441, ML383051451, ML383051521, ML383051541, ML383495341, ML488492161, ML488493071, ML488495741, ML489536041, ML490490101, ML490491211, ML492225031, ML492225721, ML492226151, ML492227061, ML494591521, ML494592781, ML612280083, ML628112445, ML628892890, ML630177372, ML630177374, ML630177375, ML631072846, ML631775027, ML80703, ML275503, ML275516, ML208763191, ML208763361, ML609054193, ML614506404, ML625284940, ML625285921, ML625285929, ML625285943, ML3783, ML136114, ML203922021, ML288889851, ML302011421, ML309991111, ML438096531, ML447814831, ML532879171, ML532879191, ML595427661, ML595427761, ML595427841, ML606000931, ML606007101, ML606007231, ML610403316, ML610403539, ML610403544, ML610403549, ML614061577, ML617436448, ML619123469, ML619123483, ML619123497, ML619123564, ML621565852, ML623549252, ML627044072, ML627044082, ML627044133, ML627044135, ML627044138, ML630815246, ML632452333, ML2973, ML203964311, ML498354061, ML588464121, ML588464141, ML588464221, ML588559161, ML588559181, ML608541575, ML609044980, ML609054864, ML609060126, ML611489476, ML611489480, ML611489570, ML616063848, ML622471409, ML624293511, ML629321005, ML71475, ML113589, ML147745061, ML147745521, ML181661171, ML182112081, ML218783941, ML318145871, ML393803991, ML438171361, ML453652031, ML453652051, ML453652061, ML453652071, ML458370601, ML527670931, ML547902961, ML552564211, ML576674691, ML609070937, ML612594787, ML615559909, ML615559910, ML615820244, ML616279077, ML616320632, ML616320633, ML616541256, ML616826623, ML617968650, ML621366811, ML632258670, ML632653461, ML632823682, ML632835335, ML612256787, ML190771, ML228265, ML295668, ML295725, ML193678891, ML281274791, ML282032271, ML482760991, ML504973261, ML622901951, ML282927, ML203903711, ML391945241, ML598702771, ML617306073, ML617853126, ML620884017, ML620884018, ML49178, ML49180, ML49195, ML58010, ML58023, ML58030, ML68057, ML139127, ML180941, ML243038, ML260575, ML260743, ML260754, ML260764, ML260765, ML264567, ML296347, ML25510561, ML25510571, ML104085391, ML104085811, ML104086021, ML104086221, ML104086381, ML104324341, ML120587851, ML165706771, ML189572931, ML203697321, ML207060361, ML207060371, ML215388241, ML226650831, ML452562921, ML461340521, ML475820671, ML475820711, ML510951321, ML518653111, ML518653511, ML594534421, ML607558741, ML608525552, ML609501570, ML609501576, ML609745927, ML611138119, ML613138384, ML615722417, ML623175063, ML623518170, ML624090793, ML624090809, ML624090818, ML624090862, ML624090882, ML624950839, ML627807110, ML631238006, ML632061713, ML128306, ML234183, ML234189, ML96363161, ML141832991, ML141833121, ML183086411, ML203489561, ML203975591, ML203990931, ML203995001, ML246157651, ML246157661, ML285643231, ML301952351, ML436955831, ML468922491, ML471428491, ML486943611, ML502544951, ML502554501, ML503166281, ML580975251, ML600338341, ML600339831, ML611425974, ML611863599, ML611863614, ML623096778, ML633105894, ML62723, ML62725, ML84870, ML84871, ML84872, ML84879, ML94451, ML112068, ML112071, ML112681, ML112682, ML112683, ML163371, ML166624, ML169007, ML169008, ML195774, ML20084291, ML27336971, ML55137931, ML57936251, ML58940631, ML99307151, ML100878661, ML102647141, ML102647161, ML102783181, ML104717361, ML148163081, ML157999611, ML158252301, ML160713181, ML161900971, ML162323001, ML163836031, ML167259671, ML167899061, ML168310631, ML168804761, ML220185621, ML221363131, ML227106451, ML232692511, ML235705721, ML238632261, ML239388461, ML240324901, ML242039281, ML243509741, ML245309111, ML245943381, ML323708501, ML325089351, ML325506631, ML330612701, ML331432011, ML339387841, ML340045421, ML340193121, ML340367651, ML341824541, ML342042601, ML342411461, ML342763941, ML343602851, ML344720451, ML345884781, ML347629171, ML349824801, ML350726861, ML352608711, ML354534641, ML357890041, ML435680681, ML447450551, ML447738361, ML449170221, ML449428811, ML450615301, ML455730761, ML457302861, ML457302871, ML460901371, ML463695891, ML464670041, ML467036281, ML467182611, ML468277491, ML468495251, ML553698981, ML556895111, ML557175071, ML567074151, ML580584281, ML615614499, ML617771625, ML618747194, ML619553267, ML619601127, ML620969579, ML135494, ML139445, ML179519, ML31727321, ML32797351, ML42547461, ML42547471, ML42547501, ML42547521, ML42547531, ML42547551, ML42823661, ML42823671, ML42823681, ML42823961, ML42824101, ML42824241, ML42824251, ML138088501, ML202767691, ML203928821, ML203928831, ML203929081, ML216442181, ML387547631, ML617130989, ML632924598, ML633096510, ML77751151, ML77751291, ML155352461, ML203974821, ML286131021, ML485580931, ML589225741, ML606424941, ML606425421, ML616546199, ML626256598, ML627496189, ML633127907, ML633127908, ML633127909, ML633127910, ML633127911, ML633127912, ML633127914, ML194116731, ML247692661, ML361863721, ML460903011, ML460903141, ML460904051, ML460906541, ML22718711, ML28124631, ML34297961, ML85341811, ML139758521, ML141721661, ML141721691, ML203909371, ML318845721, ML322097351, ML360260541, ML382260081, ML382263081, ML392199771, ML418760961, ML418890721, ML421056201, ML421056211, ML421056241, ML437327781, ML444347781, ML486215441, ML491949651, ML494668141, ML494983861, ML494983881, ML496177181, ML499619221, ML499619231, ML499619241, ML506141231, ML506141241, ML525407041, ML527671821, ML529760131, ML531570741, ML544521921, ML544521931, ML544521941, ML556898951, ML564976721, ML587540771, ML609059730, ML609603557, ML609604454, ML613931527, ML614545354, ML616332700, ML621329694, ML624533493, ML628301727, ML628340386, ML632742848, ML632742849, ML632742850, ML632742851, ML632742852, ML632742853, ML632743193, ML633055263, ML633055264, ML633055267, ML633055268, ML633055269, ML633055270, ML633055271, ML633055272, ML633055447, ML633055448, ML633055451, ML633055602, ML633055678, ML633055679, ML633055680, ML633055694, ML633055717, ML633056577, ML128990651, ML128990731, ML141325521, ML141325741, ML203887831, ML206364481, ML206364491, ML206364541, ML206736331, ML206736921, ML549913001, ML611617498, ML612241062, ML612784674, ML612784676, ML612784677, ML612949828, ML612949832, ML618706702, ML623740235, ML623740358, ML623740359, ML237481, ML237483, ML428200831, ML525939611, ML633097460, ML96394981, ML96394991, ML96395011, ML96395021, ML96395031, ML96395041, ML96545951, ML96545961, ML96545971, ML96545981, ML96545991, ML96546001, ML96546151, ML96546161, ML96546171, ML96546181, ML96546191, ML96546201, ML96546231, ML96546281, ML96546291, ML96546301, ML96546311, ML96546341, ML140036, ML241058, ML62184241, ML62184501, ML62184531, ML146540201, ML164130821, ML191521691, ML203915881, ML203953771, ML236139361, ML236273051, ML441117741, ML445308631, ML447446201, ML447447961, ML448452571, ML456783001, ML456783751, ML456785861, ML459259261, ML486954261, ML486954271, ML572056831, ML594193381, ML594193391, ML607424961, ML609784535, ML609887081, ML617217290, ML619983994, ML620213957, ML620435983, ML620739115, ML625869118, ML628856307, ML108973, ML108978, ML283410, ML283411, ML283415, ML283417, ML283419, ML283424, ML32024861, ML152499521, ML174609921, ML174610871, ML177428511, ML177428531, ML257550681, ML282152221, ML288697461, ML317628861, ML333813301, ML333813351, ML333813361, ML348671541, ML348673401, ML362134471, ML372922851, ML421352391, ML421352841, ML444724331, ML445108351, ML485268141, ML485633471, ML485789401, ML488069251, ML494743931, ML521729931, ML536007851, ML536007871, ML540403721, ML574353091, ML574436151, ML577624221, ML578719041, ML586902391, ML586904761, ML608814577, ML609070066, ML609070717, ML609070718, ML610162055, ML610162056, ML611257310, ML611719602, ML611960225, ML612335521, ML612335522, ML614336063, ML614336287, ML614336314, ML614336315, ML614336316, ML615630806, ML616402951, ML617169305, ML625195065, ML627264261, ML627265605, ML627454140, ML627454141, ML628028879, ML630843276, ML630843277, ML630843278, ML631514717, ML631514718, ML631514719, ML631605423, ML632646742, ML632646743, ML632646744, ML633103465, ML106440, ML203971931, ML204019821, ML81101, ML117256, ML117257, ML117258, ML117259, ML126757, ML126906, ML172571, ML173790, ML178427, ML178431, ML178432, ML199608, ML202300, ML222300, ML222345, ML229333, ML41193631, ML42518481, ML45601911, ML52634301, ML76070881, ML76072141, ML76072161, ML76072241, ML76072261, ML106825751, ML110898061, ML171147541, ML176922471, ML176929541, ML189798981, ML194645771, ML203701211, ML203989301, ML206794651, ML207103961, ML217507871, ML230257281, ML230257961, ML230258861, ML242214261, ML250340111, ML259130141, ML268430871, ML294441271, ML301788301, ML301788311, ML301788321, ML305972241, ML321845211, ML364638471, ML364638711, ML366708501, ML366708511, ML366708521, ML396567931, ML421098201, ML421629831, ML432271111, ML441023581, ML445052161, ML445052171, ML461610641, ML537312391, ML540698681, ML548974301, ML549690721, ML556041261, ML573461591, ML585042071, ML593165761, ML602345661, ML603609061, ML603609071, ML607878831, ML609062217, ML609070655, ML609070676, ML610412050, ML610474319, ML615081421, ML615182051, ML615182056, ML615251614, ML615704261, ML616148791, ML616517701, ML616517703, ML616822240, ML616835654, ML616928978, ML617206666, ML617259105, ML617303761, ML617418434, ML617788689, ML619429515, ML619599797, ML620333302, ML621800808, ML621800809, ML621808490, ML621810005, ML621810135, ML621938484, ML623034712, ML623034728, ML623084388, ML625733287, ML627874744, ML629641128, ML630666435, ML631206357, ML631338902, ML631339113, ML631495795, ML631657086, ML631657093, ML631895053, ML631895751, ML632741568, ML632741569, ML632741570, ML632741571, ML632741572, ML632741732, ML632742026, ML500, ML501, ML282194, ML40149381, ML48312121, ML66265691, ML89582541, ML95857851, ML105698151, ML115333631, ML119390741, ML119390831, ML125611911, ML139426501, ML146317281, ML153072731, ML178504801, ML179835091, ML179835131, ML218916061, ML218917051, ML218919861, ML229541261, ML232086561, ML242130011, ML246800891, ML257580861, ML257581681, ML263860151, ML263860161, ML275470111, ML277470461, ML279851511, ML279851541, ML285082671, ML288173151, ML288173161, ML303879741, ML332825291, ML355878201, ML363347771, ML366419331, ML385408531, ML387586281, ML395280711, ML395827851, ML398248971, ML416699031, ML416930131, ML446090411, ML469363631, ML483495611, ML485511721, ML492293801, ML492293811, ML529430611, ML530697781, ML539577771, ML539577781, ML539577791, ML542040151, ML553810521, ML565994981, ML579834511, ML579893811, ML581428491, ML581428501, ML587723391, ML587723401, ML587723411, ML596281361, ML602993441, ML609497536, ML611189505, ML612324796, ML613969957, ML614629694, ML614629762, ML614629763, ML614629764, ML614629765, ML614806221, ML615662465, ML615662515, ML618897947, ML619981472, ML620171766, ML621964163, ML624017786, ML625425849, ML627085632, ML628717371, ML629646126, ML629689915, ML630120748, ML630273621, ML630273841, ML632091164, ML632820430, ML633103550, ML180925, ML245680, ML43562281, ML203933941, ML207628671, ML151625201, ML151625211, ML151625231, ML151625241, ML151625251, ML151625271, ML151625301, ML151625311, ML239479211, ML248957471, ML253565131, ML253565521, ML253565911, ML253566281, ML253566491, ML253567381, ML253568771, ML317391501, ML123753451, ML196658021, ML203951971, ML211581051, ML278557821, ML442888471, ML442888731, ML453711581, ML491361501, ML491361511, ML491361531, ML491361541, ML491361631, ML592393141, ML592393151, ML609055864, ML609055865, ML611535407, ML611535577, ML611535653, ML614481045, ML614481046, ML618496554, ML619941176, ML619941786, ML619941887, ML619941888, ML619942209, ML619942210, ML620029809, ML620029810, ML620029811, ML620029813, ML620029814, ML620030242, ML620162627, ML630265814, ML630273410, ML630273575, ML630784297, ML84004, ML135777, ML137673, ML137674, ML137676, ML137681, ML139400, ML139403, ML139407, ML139408, ML139409, ML139414, ML144652, ML215881, ML238159, ML242021, ML58633841, ML58633861, ML77478421, ML134627021, ML138351591, ML140940991, ML165655801, ML165658131, ML177368311, ML182658181, ML203960831, ML206475451, ML210267271, ML213773931, ML216965411, ML248359931, ML281856681, ML325529581, ML372183591, ML372183601, ML372183791, ML372186531, ML439106751, ML439701101, ML439701141, ML445367061, ML445367141, ML451356181, ML453950751, ML454311321, ML487020871, ML516806061, ML524608381, ML524608391, ML534536881, ML534537711, ML591776431, ML602147041, ML607783661, ML625643420, ML627918251, ML630406410, ML630406411, ML31676241, ML203887851, ML290065611, ML290065651, ML290065711, ML290065721, ML290065991, ML411736801, ML451851581, ML454758041, ML29876, ML29948, ML58944, ML58945, ML58946, ML58947, ML58948, ML141326, ML184953, ML187905, ML220191, ML235734, ML112852841, ML112852851, ML112852861, ML112852871, ML125818921, ML144560411, ML189879851, ML189880381, ML219097311, ML234713821, ML306736691, ML396511701, ML407856871, ML410618161, ML503936451, ML520718451, ML568846691, ML569490561, ML569492871, ML569494461, ML569502761, ML569505271, ML618952460, ML625380540, ML629035266, ML24008041, ML203926781, ML364488391, ML613918054, ML613918106, ML613918153, ML613918208, ML631712242, ML631712243, ML31370, ML41240, ML43506, ML43509, ML49111, ML49442, ML61385, ML63458, ML81191, ML81192, ML81196, ML83890, ML84068, ML84281, ML84282, ML85699, ML86021, ML139083, ML191966, ML242954, ML242976, ML242998, ML243940, ML244463, ML244465, ML260554, ML260577, ML25509921, ML25510791, ML103962331, ML103962871, ML103962921, ML142070491, ML142077031, ML144668811, ML164255501, ML166543861, ML167088071, ML178057351, ML178057611, ML179475751, ML203699241, ML213019591, ML213630171, ML213644361, ML216790611, ML216875311, ML221209381, ML222608181, ML234937951, ML284515831, ML289218811, ML314928061, ML317316751, ML351139791, ML358264971, ML360201001, ML375889531, ML376108921, ML475832171, ML486794091, ML492899511, ML492899531, ML494565561, ML494709971, ML494710581, ML503683771, ML509522141, ML509522151, ML509522161, ML509529661, ML509529671, ML509529681, ML509529691, ML509529721, ML519972161, ML520112891, ML522329351, ML527753981, ML533603961, ML605472431, ML605472451, ML608863664, ML609305501, ML609511946, ML610383238, ML612347508, ML615447029, ML615962782, ML616620392, ML618631309, ML621188737, ML621902208, ML622336536, ML626196869, ML628264597, ML628986478, ML630283658, ML632063532, ML632063533, ML43431, ML53818, ML53832, ML69500, ML129730, ML129731, ML188065, ML26248821, ML79585241, ML91655331, ML199646151, ML204013521, ML304070371, ML352879881, ML352880281, ML352889751, ML361006821, ML361006831, ML421458831, ML463729791, ML464330041, ML468723851, ML483315711, ML483316201, ML506332541, ML536749561, ML540460031, ML541308561, ML562413951, ML572966461, ML574818531, ML583841841, ML594345141, ML609056825, ML609057521, ML613123184, ML615881106, ML615896393, ML616408543, ML616408552, ML616408831, ML616408832, ML617133446, ML620954424, ML621277635, ML622200181, ML625669638, ML626054222, ML626054299, ML627048579, ML627660297, ML629669158, ML630012535, ML630071648, ML630791578, ML631040549, ML631922855, ML631922860, ML632343313, ML632364343, ML632544882, ML632652349, ML632652350, ML632778952, ML163663, ML163667, ML163710, ML163713, ML163714, ML42751381, ML90737921, ML90737981, ML90780421, ML90916111, ML90918211, ML181148831, ML181148881, ML203898461, ML340148111, ML360222331, ML528725331, ML533810361, ML608805281, ML608805295, ML608912254, ML608988516, ML608988554, ML610459771, ML610459772, ML612749476, ML612749487, ML613170796, ML613170797, ML613170798, ML613170799, ML622540341, ML622834259, ML623353495, ML623353496, ML624643173, ML630032916, ML630032933, ML85012701, ML203969961, ML556781501, ML588887511, ML616038861, ML617126643, ML621068453, ML39814, ML145838, ML145839, ML164504, ML236426, ML151358811, ML203901221, ML221171401, ML222238001, ML222238011, ML224974081, ML282448931, ML323489261, ML337277981, ML440325631, ML453373531, ML453373561, ML453373601, ML453377261, ML460282481, ML522143121, ML522143131, ML522143141, ML536015261, ML536015271, ML536015281, ML578182921, ML608461309, ML608461310, ML608461311, ML608461312, ML611126718, ML612782559, ML612782560, ML614337128, ML616536268, ML616817769, ML618196934, ML624782666, ML624783009, ML627261377, ML627261380, ML627261382, ML628472601, ML629899457, ML632746585, ML518112, ML518186, ML518208, ML175421341, ML218135721, ML380589071, ML466942521, ML578221721, ML609065097, ML609069603, ML616197170, ML621044592, ML627491184, ML627491185, ML627818482, ML629670585, ML629670589, ML629670590, ML629670591, ML629670592, ML630439435, ML632630109, ML101998, ML566335691, ML566335751, ML566335801, ML566335871, ML566336051, ML566341071, ML618707335, ML609060533, ML609060536, ML298236081, ML613865949, ML613865958, ML613865961, ML613865971, ML622991043, ML626720751, ML629078350, ML631031074, ML631031130, ML631031131, ML631031167, ML631031225, ML631031319, ML631031320, ML631031321, ML620518317, ML620737788, ML620737952, ML95421, ML20380691, ML203930221, ML611037516, ML611037746, ML624431435, ML626248324, ML135486, ML135488, ML163290, ML163291, ML31729111, ML616277005, ML616908429, ML93646, ML93649, ML95981, ML95987, ML95989, ML95994, ML100022, ML204006871, ML591296471, ML125923, ML210653, ML539516, ML34213541, ML34213651, ML73082721, ML73082761, ML100610221, ML116212531, ML124208581, ML125511641, ML141341811, ML152674091, ML156104021, ML169925401, ML170111681, ML171248371, ML181966621, ML181966661, ML195894971, ML195895011, ML195895151, ML195895261, ML195898081, ML203993241, ML203993251, ML245537801, ML259607971, ML263108341, ML263316141, ML263318461, ML263318511, ML263318711, ML266649721, ML266652291, ML266652351, ML266712871, ML268901391, ML268901551, ML272345111, ML330029281, ML330477251, ML330477261, ML331967731, ML332195441, ML332747981, ML332747991, ML341095441, ML351069941, ML361812621, ML362334371, ML363846091, ML363846421, ML369829121, ML373437101, ML374000121, ML396697101, ML446933651, ML459008031, ML462559101, ML462561221, ML462561241, ML469290021, ML469290051, ML470884341, ML472185361, ML474308791, ML478672151, ML479851811, ML479851821, ML480324481, ML488443751, ML570614511, ML574078541, ML580585711, ML580585911, ML590618261, ML590624121, ML590919231, ML611569397, ML612976472, ML612976473, ML612976474, ML613595383, ML613595384, ML614329746, ML615201508, ML615201513, ML617462844, ML619268540, ML620475385, ML620475387, ML620521117, ML620629466, ML621268948, ML621622963, ML622157248, ML625388112, ML627108325, ML631267395, ML88144, ML154660821, ML181669941, ML204002531, ML227715051, ML258420041, ML338813091, ML404393911, ML486937221, ML495238401, ML503599951, ML503599961, ML503599971, ML507319441, ML522480761, ML522782051, ML522834681, ML533520571, ML538275621, ML609057884, ML615278430, ML615278484, ML615575840, ML615665429, ML615666580, ML618413462, ML620946300, ML623829678, ML627419956, ML630778458, ML632083567, ML632742848, ML632742849, ML632742850, ML632742851, ML632742852, ML632742853, ML632743193, ML633055263, ML633055264, ML633055267, ML633055268, ML633055269, ML633055270, ML633055271, ML633055272, ML633055447, ML633055448, ML633055451, ML633055602, ML633055678, ML633055679, ML633055680, ML633055694, ML633055717, ML633056577, ML128990651, ML128990731, ML141325521, ML141325741, ML203887831, ML206364481, ML206364491, ML206364541, ML206736331, ML206736921, ML549913001, ML611617498, ML612241062, ML612784674, ML612784676, ML612784677, ML612949828, ML612949832, ML618706702, ML623740235, ML623740358, ML623740359, ML237481, ML237483, ML428200831, ML525939611, ML633097460, ML96394981, ML96394991, ML96395011, ML96395021, ML96395031, ML96395041, ML96545951, ML96545961, ML96545971, ML96545981, ML96545991, ML96546001, ML96546151, ML96546161, ML96546171, ML96546181, ML96546191, ML96546201, ML96546231, ML96546281, ML96546291, ML96546301, ML96546311, ML96546341, ML140036, ML241058, ML62184241, ML62184501, ML62184531, ML146540201, ML164130821, ML191521691, ML203915881, ML203953771, ML236139361, ML236273051, ML441117741, ML445308631, ML447446201, ML447447961, ML448452571, ML456783001, ML456783751, ML456785861, ML459259261, ML486954261, ML486954271, ML572056831, ML594193381, ML594193391, ML607424961, ML609784535, ML609887081, ML617217290, ML619983994, ML620213957, ML620435983, ML620739115, ML625869118, ML628856307, ML108973, ML108978, ML283410, ML283411, ML283415, ML283417, ML283419, ML283424, ML32024861, ML152499521, ML174609921, ML174610871, ML177428511, ML177428531, ML257550681, ML282152221, ML288697461, ML317628861, ML333813301, ML333813351, ML333813361, ML348671541, ML348673401, ML362134471, ML372922851, ML421352391, ML421352841, ML444724331, ML445108351, ML485268141, ML485633471, ML485789401, ML488069251, ML494743931, ML521729931, ML536007851, ML536007871, ML540403721, ML574353091, ML574436151, ML577624221, ML578719041, ML586902391, ML586904761, ML608814577, ML609070066, ML609070717, ML609070718, ML610162055, ML610162056, ML611257310, ML611719602, ML611960225, ML612335521, ML612335522, ML614336063, ML614336287, ML614336314, ML614336315, ML614336316, ML615630806, ML616402951, ML617169305, ML625195065, ML627264261, ML627265605, ML627454140, ML627454141, ML628028879, ML630843276, ML630843277, ML630843278, ML631514717, ML631514718, ML631514719, ML631605423, ML632646742, ML632646743, ML632646744, ML633103465, ML106440, ML203971931, ML204019821, ML81101, ML117256, ML117257, ML117258, ML117259, ML126757, ML126906, ML172571, ML173790, ML178427, ML178431, ML178432, ML199608, ML202300, ML222300, ML222345, ML229333, ML41193631, ML42518481, ML45601911, ML52634301, ML76070881, ML76072141, ML76072161, ML76072241, ML76072261, ML106825751, ML110898061, ML171147541, ML176922471, ML176929541, ML189798981, ML194645771, ML203701211, ML203989301, ML206794651, ML207103961, ML217507871, ML230257281, ML230257961, ML230258861, ML242214261, ML250340111, ML259130141, ML268430871, ML294441271, ML301788301, ML301788311, ML301788321, ML305972241, ML321845211, ML364638471, ML364638711, ML366708501, ML366708511, ML366708521, ML396567931, ML421098201, ML421629831, ML432271111, ML441023581, ML445052161, ML445052171, ML461610641, ML537312391, ML540698681, ML548974301, ML549690721, ML556041261, ML573461591, ML585042071, ML593165761, ML602345661, ML603609061, ML603609071, ML607878831, ML609062217, ML609070655, ML609070676, ML610412050, ML610474319, ML615081421, ML615182051, ML615182056, ML615251614, ML615704261, ML616148791, ML616517701, ML616517703, ML616822240, ML616835654, ML616928978, ML617206666, ML617259105, ML617303761, ML617418434, ML617788689, ML619429515, ML619599797, ML620333302, ML621800808, ML621800809, ML621808490, ML621810005, ML621810135, ML621938484, ML623034712, ML623034728, ML623084388, ML625733287, ML627874744, ML629641128, ML630666435, ML631206357, ML631338902, ML631339113, ML631495795, ML631657086, ML631657093, ML631895053, ML631895751, ML632741568, ML632741569, ML632741570, ML632741571, ML632741572, ML632741732, ML632742026, ML500, ML501, ML282194, ML40149381, ML48312121, ML66265691, ML89582541, ML95857851, ML105698151, ML115333631, ML119390741, ML119390831, ML125611911, ML139426501, ML146317281, ML153072731, ML178504801, ML179835091, ML179835131, ML218916061, ML218917051, ML218919861, ML229541261, ML232086561, ML242130011, ML246800891, ML257580861, ML257581681, ML263860151, ML263860161, ML275470111, ML277470461, ML279851511, ML279851541, ML285082671, ML288173151, ML288173161, ML303879741, ML332825291, ML355878201, ML363347771, ML366419331, ML385408531, ML387586281, ML395280711, ML395827851, ML398248971, ML416699031, ML416930131, ML446090411, ML469363631, ML483495611, ML485511721, ML492293801, ML492293811, ML529430611, ML530697781, ML539577771, ML539577781, ML539577791, ML542040151, ML553810521, ML565994981, ML579834511, ML579893811, ML581428491, ML581428501, ML587723391, ML587723401, ML587723411, ML596281361, ML602993441, ML609497536, ML611189505, ML612324796, ML613969957, ML614629694, ML614629762, ML614629763, ML614629764, ML614629765, ML614806221, ML615662465, ML615662515, ML618897947, ML619981472, ML620171766, ML621964163, ML624017786, ML625425849, ML627085632, ML628717371, ML629646126, ML629689915, ML630120748, ML630273621, ML630273841, ML632091164, ML632820430, ML633103550, ML180925, ML245680, ML43562281, ML203933941, ML207628671, ML151625201, ML151625211, ML151625231, ML151625241, ML151625251, ML151625271, ML151625301, ML151625311, ML239479211, ML248957471, ML253565131, ML253565521, ML253565911, ML253566281, ML253566491, ML253567381, ML253568771, ML317391501, ML123753451, ML196658021, ML203951971, ML211581051, ML278557821, ML442888471, ML442888731, ML453711581, ML491361501, ML491361511, ML491361531, ML491361541, ML491361631, ML592393141, ML592393151, ML609055864, ML609055865, ML611535407, ML611535577, ML611535653, ML614481045, ML614481046, ML618496554, ML619941176, ML619941786, ML619941887, ML619941888, ML619942209, ML619942210, ML620029809, ML620029810, ML620029811, ML620029813, ML620029814, ML620030242, ML620162627, ML630265814, ML630273410, ML630273575, ML630784297, ML84004, ML135777, ML137673, ML137674, ML137676, ML137681, ML139400, ML139403, ML139407, ML139408, ML139409, ML139414, ML144652, ML215881, ML238159, ML242021, ML58633841, ML58633861, ML77478421, ML134627021, ML138351591, ML140940991, ML165655801, ML165658131, ML177368311, ML182658181, ML203960831, ML206475451, ML210267271, ML213773931, ML216965411, ML248359931, ML281856681, ML325529581, ML372183591, ML372183601, ML372183791, ML372186531, ML439106751, ML439701101, ML439701141, ML445367061, ML445367141, ML451356181, ML453950751, ML454311321, ML487020871, ML516806061, ML524608381, ML524608391, ML534536881, ML534537711, ML591776431, ML602147041, ML607783661, ML625643420, ML627918251, ML630406410, ML630406411, ML31676241, ML203887851, ML290065611, ML290065651, ML290065711, ML290065721, ML290065991, ML411736801, ML451851581, ML454758041, ML29876, ML29948, ML58944, ML58945, ML58946, ML58947, ML58948, ML141326, ML184953, ML187905, ML220191, ML235734, ML112852841, ML112852851, ML112852861, ML112852871, ML125818921, ML144560411, ML189879851, ML189880381, ML219097311, ML234713821, ML306736691, ML396511701, ML407856871, ML410618161, ML503936451, ML520718451, ML568846691, ML569490561, ML569492871, ML569494461, ML569502761, ML569505271, ML618952460, ML625380540, ML629035266, ML24008041, ML203926781, ML364488391, ML613918054, ML613918106, ML613918153, ML613918208, ML631712242, ML631712243, ML31370, ML41240, ML43506, ML43509, ML49111, ML49442, ML61385, ML63458, ML81191, ML81192, ML81196, ML83890, ML84068, ML84281, ML84282, ML85699, ML86021, ML139083, ML191966, ML242954, ML242976, ML242998, ML243940, ML244463, ML244465, ML260554, ML260577, ML25509921, ML25510571, ML103962331, ML103962871, ML103962921, ML142070491, ML142077031, ML144668811, ML164255501, ML166543861, ML167088071, ML178057351, ML178057611, ML179475751, ML203699241, ML213019591, ML213630171, ML213644361, ML216790611, ML216875311, ML221209381, ML222608181, ML234937951, ML284515831, ML289218811, ML314928061, ML317316751, ML351139791, ML358264971, ML360201001, ML375889531, ML376108921, ML475832171, ML486794091, ML492899511, ML492899531, ML494565561, ML494709971, ML494710581, ML503683771, ML509522141, ML509522151, ML509522161, ML509529661, ML509529671, ML509529681, ML509529691, ML509529721, ML519972161, ML520112891, ML522329351, ML527753981, ML533603961, ML605472431, ML605472451, ML608863664, ML609305501, ML609511946, ML610383238, ML612347508, ML615447029, ML615962782, ML616620392, ML618631309, ML621188737, ML621902208, ML622336536, ML626196869, ML628264597, ML628986478, ML630283658, ML632063532, ML632063533, ML43431, ML53818, ML53832, ML69500, ML129730, ML129731, ML188065, ML26248821, ML79585241, ML91655331, ML199646151, ML204013521, ML304070371, ML352879881, ML352880281, ML352889751, ML361006821, ML361006831, ML421458831, ML463729791, ML464330041, ML468723851, ML483315711, ML483316201, ML506332541, ML536749561, ML540460031, ML541308561, ML562413951, ML572966461, ML574818531, ML583841841, ML594345141, ML609056825, ML609057521, ML613123184, ML615881106, ML615896393, ML616408543, ML616408552, ML616408831, ML616408832, ML617133446, ML620954424, ML621277635, ML622200181, ML624017786, ML625425849, ML627085632, ML628717371, ML629646126, ML629689915, ML630120748, ML630273621, ML630273841, ML632091164, ML632820430, ML633103550, ML180925, ML245680, ML43562281, ML203933941, ML207628671, ML151625201, ML151625211, ML151625231, ML151625241, ML151625251, ML151625271, ML151625301, ML151625311, ML239479211, ML248957471, ML253565131, ML253565521, ML253565911, ML253566281, ML253566491, ML253567381, ML253568771, ML317391501, ML123753451, ML196658021, ML203951971, ML211581051, ML278557821, ML442888471, ML442888731, ML453711581, ML491361501, ML491361511, ML491361531, ML491361541, ML491361631, ML592393141, ML592393151, ML609055864, ML609055865, ML611535407, ML611535577, ML611535653, ML614481045, ML614481046, ML618496554, ML619941176, ML619941786, ML619941887, ML619941888, ML619942209, ML619942210, ML620029809, ML620029810, ML620029811, ML620029813, ML620029814, ML620030242, ML620162627, ML630265814, ML630273410, ML630273575, ML630784297, ML84004, ML135777, ML137673, ML137674, ML137676, ML137681, ML139400, ML139403, ML139407, ML139408, ML139409, ML139414, ML144652, ML215881, ML238159, ML242021, ML58633841, ML58633861, ML77478421, ML134627021, ML138351591, ML140940991, ML165655801, ML165658131, ML177368311, ML182658181, ML203960831, ML206475451, ML210267271, ML213773931, ML216965411, ML248359931, ML281856681, ML325529581, ML372183591, ML372183601, ML372183791, ML372186531, ML439106751, ML439701101, ML439701141, ML445367061, ML445367141, ML451356181, ML453950751, ML454311321, ML487020871, ML516806061, ML524608381, ML524608391, ML534536881, ML534537711, ML591776431, ML602147041, ML607783661, ML625643420, ML627918251, ML630406410, ML630406411, ML31676241, ML203887851, ML290065611, ML290065651, ML290065711, ML290065721, ML290065991, ML411736801, ML451851581, ML454758041, ML29876, ML29948, ML58944, ML58945, ML58946, ML58947, ML58948, ML141326, ML184953, ML187905, ML220191, ML235734, ML112852841, ML112852851, ML112852861, ML112852871, ML125818921, ML144560411, ML189879851, ML189880381, ML219097311, ML234713821, ML306736691, ML396511701, ML407856871, ML410618161, ML503936451, ML520718451, ML568846691, ML569490561, ML569492871, ML569494461, ML569502761, ML569505271, ML618952460, ML625380540, ML629035266, ML24008041, ML203926781, ML364488391, ML613918054, ML613918106, ML613918153, ML613918208, ML631712242, ML631712243, ML31370, ML41240, ML43506, ML43509, ML49111, ML49442, ML61385, ML63458, ML81191, ML81192, ML81196, ML83890, ML84068, ML84281, ML84282, ML85699, ML86021, ML139083, ML191966, ML242954, ML242976, ML242998, ML243940, ML244463, ML244465, ML260554, ML260577, ML25509921, ML25510791, ML103962331, ML103962871, ML103962921, ML142070491, ML142077031, ML144668811, ML164255501, ML166543861, ML167088071, ML178057351, ML178057611, ML179475751, ML203699241, ML213019591, ML213630171, ML213644361, ML216790611, ML216875311, ML221209381, ML222608181, ML234937951, ML284515831, ML289218811, ML314928061, ML317316751, ML351139791, ML358264971, ML360201001, ML375889531, ML376108921, ML475832171, ML486794091, ML492899511, ML492899531, ML494565561, ML494709971, ML494710581, ML503683771, ML509522141, ML509522151, ML509522161, ML509529661, ML509529671, ML509529681, ML509529691, ML509529721, ML519972161, ML520112891, ML522329351, ML527753981, ML533603961, ML605472431, ML605472451, ML608863664, ML609305501, ML609511946, ML610383238, ML612347508, ML615447029, ML615962782, ML616620392, ML618631309, ML621188737, ML621902208, ML622336536, ML626196869, ML628264597, ML628986478, ML630283658, ML632063532, ML632063533, ML43431, ML53818, ML53832, ML69500, ML129730, ML129731, ML188065, ML26248821, ML79585241, ML91655331, ML199646151, ML204013521, ML304070371, ML352879881, ML352880281, ML352889751, ML361006821, ML361006831, ML421458831, ML463729791, ML464330041, ML468723851, ML483315711, ML483316201, ML506332541, ML536749561, ML540460031, ML541308561, ML562413951, ML572966461, ML574818531, ML583841841, ML594345141, ML609056825, ML609057521, ML613123184, ML615881106, ML615896393, ML616408543, ML616408552, ML616408831, ML616408832, ML617133446, ML620954424, ML621277635, ML622200181, ML625669638, ML626054222, ML626054299, ML627048579, ML627660297, ML629669158, ML630012535, ML630071648, ML630791578, ML631040549, ML631922855, ML631922860, ML632343313, ML632364343, ML632544882, ML632652349, ML632652350, ML632778952, ML163663, ML163667, ML163710, ML163713, ML163714, ML42751381, ML90737921, ML90737981, ML90780421, ML90916111, ML90918211, ML181148831, ML181148881, ML203898461, ML340148111, ML360222331, ML528725331, ML533810361, ML608805281, ML608805295, ML608912254, ML608988516, ML608988554, ML610459771, ML610459772, ML612749476, ML612749487, ML613170796, ML613170797, ML613170798, ML613170799, ML622540341, ML622834259, ML623353495, ML623353496, ML624643173, ML630032916, ML630032933, ML85012701, ML203969961, ML556781501, ML588887511, ML616038861, ML617126643, ML621068453, ML39814, ML145838, ML145839, ML164504, ML236426, ML151358811, ML203901221, ML221171401, ML222238001, ML222238011, ML224974081, ML282448931, ML323489261, ML337277981, ML440325631, ML453373531, ML453373561, ML453373601, ML453377261, ML460282481, ML522143121, ML522143131, ML522143141, ML536015261, ML536015271, ML536015281, ML578182921, ML608461309, ML608461310, ML608461311, ML608461312, ML611126718, ML612782559, ML612782560, ML614337128, ML616536268, ML616817769, ML618196934, ML624782666, ML624783009, ML627261377, ML627261380, ML627261382, ML628472601, ML629899457, ML632746585, ML518112, ML518186, ML518208, ML175421341, ML218135721, ML380589071, ML466942521, ML578221721, ML609065097, ML609069603, ML616197170, ML621044592, ML627491184, ML627491185, ML627818482, ML629670585, ML629670589, ML629670590, ML629670591, ML629670592, ML630439435, ML632630109, ML101998, ML566335691, ML566335751, ML566335801, ML566335871, ML566336051, ML566341071, ML618707335, ML609060533, ML609060536, ML298236081, ML613865949, ML613865958, ML613865961, ML622991043, ML626720751, ML629078350, ML631031074, ML631031130, ML631031131, ML631031167, ML631031225, ML631031319, ML631031320, ML631031321, ML620518317, ML620737788, ML620737952, ML95421, ML20380691, ML203930221, ML611037516, ML611037746, ML624431435, ML626248324, ML135486, ML135488, ML163290, ML163291, ML31729111, ML616277005, ML616908429, ML93646, ML93649, ML95981, ML95987, ML95989, ML95994, ML100022, ML204006871, ML591296471, ML125923, ML210653, ML539516, ML34213541, ML34213651, ML73082721, ML73082761, ML100610221, ML116212531, ML124208581, ML125511641, ML141341811, ML152674091, ML156104021, ML169925401, ML170111681, ML171248371, ML181966621, ML181966661, ML195894971, ML195895011, ML195895151, ML195895261, ML195898081, ML203993241, ML203993251, ML245537801, ML259607971, ML263108341, ML263316141, ML263318461, ML263318511, ML263318711, ML266649721, ML266652291, ML266652351, ML266712871, ML268901391, ML268901551, ML272345111, ML330029281, ML330477251, ML330477261, ML331967731, ML332195441, ML332747981, ML332747991, ML341095441, ML351069941, ML361812621, ML362334371, ML363846091, ML363846421, ML369829121, ML373437101, ML374000121, ML396697101, ML446933651, ML459008031, ML462559101, ML462561221, ML462561241, ML469290021, ML469290051, ML470884341, ML472185361, ML474308791, ML478672151, ML479851811, ML479851821, ML480324481, ML488443751, ML570614511, ML574078541, ML580585711, ML580585911, ML590618261, ML590624121, ML590919231, ML611569397, ML612976472, ML612976473, ML612976474, ML613595383, ML613595384, ML614329746, ML615201508, ML615201513, ML617462844, ML619268540, ML620475385, ML620475387, ML620521117, ML620629466, ML621268948, ML621622963, ML622157248, ML625388112, ML627108325, ML631267395, ML88144, ML154660821, ML181669941, ML204002531, ML227715051, ML258420041, ML338813091, ML404393911, ML486937221, ML495238401, ML503599951, ML503599961, ML503599971, ML507319441, ML522480761, ML522782051, ML522834681, ML533520571, ML538275621, ML609057884, ML615278430, ML615278484, ML615575840, ML615665429, ML615666580, ML618413462, ML620946300, ML623829678, ML627419956, ML630778458, ML632083567, ML55407, ML147856, ML216502, ML216520, ML216523, ML216526, ML216530, ML216532, ML216533, ML216537, ML164940031, ML164940041, ML203988651, ML261407841, ML346502371, ML356159651, ML356159931, ML356160841, ML356167401, ML359997581, ML365551911, ML365559051, ML474169771, ML608580853, ML621284345, ML622718822, ML622804536, ML622804550, ML622804559, ML47878591, ML89632391, ML599088831, ML599095091, ML619307334, ML628304869, ML633037338, ML55895, ML63639, ML63801, ML132936, ML135833, ML135834, ML139228, ML139229, ML139231, ML178062, ML178176, ML199164, ML199193, ML295861, ML295870, ML295872, ML295875, ML60836051, ML71674251, ML109119571, ML173406981, ML182617711, ML182617721, ML187438631, ML203960831, ML204008341, ML209873211, ML221764091, ML223365571, ML224427061, ML272193901, ML273042831, ML273045891, ML275485031, ML286003581, ML289218641, ML301001431, ML301818731, ML304578371, ML310811611, ML326918711, ML337159511, ML350177411, ML356102731, ML357210981, ML357211441, ML370847811, ML434928551, ML448350601, ML450450301, ML463108081, ML512220291, ML518535821, ML523186251, ML534925441, ML540481621, ML557749111, ML578741001, ML596848761, ML603051691, ML603128571, ML608278156, ML608471376, ML609271908, ML609301953, ML610073145, ML610073146, ML610361988, ML612230239, ML612467305, ML612594289, ML614147950, ML614696429, ML615430281, ML618282554, ML618282939, ML618284836, ML618285166, ML622138326, ML622854758, ML623256708, ML623267671, ML624860512, ML626036391, ML626036589, ML626036601, ML626060783, ML626060784, ML626130900, ML626308869, ML626308870, ML626317502, ML626334662, ML626364605, ML628209827, ML628807080, ML628807081, ML629143991, ML630241567, ML630401361, ML28663, ML28667, ML28668, ML28670, ML28673, ML28676, ML45973, ML45974, ML45975, ML45976, ML45977, ML45978, ML45979, ML45980, ML45981, ML45982, ML45983, ML45984, ML46240, ML46241, ML46242, ML46243, ML46244, ML46245, ML46246, ML46247, ML46248, ML46249, ML46250, ML82549, ML82550, ML82551, ML82552, ML82553, ML82554, ML82555, ML82556, ML82557, ML82558, ML82559, ML82560, ML82561, ML82562, ML82563, ML82564, ML82854, ML82855, ML82856, ML82857, ML82858, ML82859, ML82860, ML82861, ML82862, ML86735, ML86736, ML86737, ML86738, ML86739, ML86740, ML86741, ML86742, ML86743, ML86744, ML86745, ML86746, ML86747, ML86748, ML86749, ML86750, ML86751, ML86752, ML86780, ML237478, ML82697421, ML133679211, ML133679301, ML133680871, ML203698311, ML457686091, ML461771471, ML461775151, ML462884781, ML496647551, ML525932501, ML612226811, ML626444431, ML218463, ML310397, ML167173401, ML172342661, ML203698321, ML620926799, ML620928496, ML620968030, ML620968037, ML621989983, ML71301, ML71302, ML164124, ML164139, ML164192, ML183402, ML213522, ML213526, ML282817, ML282843, ML282850, ML282852, ML86809761, ML203998351, ML203998361, ML221387011, ML468231121, ML497257831, ML549652121, ML558464561, ML577444281, ML609065059, ML609065060, ML611226846, ML611226847, ML611226848, ML615208725, ML616148637, ML616511772, ML616511773, ML616821203, ML616821204, ML619007035, ML619438754, ML620372305, ML620372308, ML621128672, ML621165512, ML621196138, ML621780700, ML621780701, ML621780702, ML622189372, ML626123294, ML626123326, ML626646907, ML628411552, ML628411556, ML629219870, ML629219886, ML629908020, ML631112603, ML974, ML975, ML976, ML102178, ML203976201, ML203976211, ML203976221, ML204018371, ML392875191, ML393879251, ML414693841, ML414694771, ML417616501, ML417617911, ML417621601, ML488348921, ML532805421, ML532805431, ML593564181, ML614011508, ML616423298, ML617299458, ML617299459, ML617299460, ML623217084, ML623217113, ML17388, ML18076, ML26914, ML28741, ML35907, ML58885, ML78982, ML165289, ML185631, ML203413, ML308362, ML537443, ML24042311, ML65697691, ML99101011, ML101138051, ML101138441, ML101138661, ML106360361, ML106360371, ML106374991, ML107358191, ML123343561, ML126817791, ML127815511, ML129649461, ML167164091, ML181425051, ML181425061, ML181426281, ML203697311, ML203915711, ML218280101, ML221712241, ML224056191, ML224115871, ML233689241, ML251695321, ML311688031, ML314109891, ML314827381, ML324628081, ML330452941, ML353016651, ML372099791, ML391718331, ML394538871, ML395439871, ML404591651, ML420105241, ML421780441, ML430978591, ML442739651, ML443048451, ML469440241, ML480381981, ML505272371, ML512892711, ML524428081, ML528651431, ML530544881, ML535078431, ML562177331, ML562177341, ML594845601, ML604328411, ML608558586, ML609330676, ML609330687, ML609777470, ML609777471, ML609777476, ML609777477, ML610979565, ML611515603, ML613336671, ML613976202, ML614033302, ML614033303, ML614033304, ML614265134, ML616072441, ML616895005, ML616935890, ML619035395, ML619746997, ML621177061, ML621177291, ML621177365, ML621177649, ML621941679, ML621995926, ML622180988, ML622862224, ML626156140, ML626156141, ML627113338, ML630299883, ML632346770, ML632347070, ML6124, ML44285, ML64918, ML286413, ML286426, ML143601051, ML228270331, ML299902461, ML309014751, ML387941341, ML390117271, ML390117281, ML323642761, ML384659661, ML36908, ML70457, ML71208, ML71271, ML71274, ML71286, ML71287, ML71335, ML71351, ML71485, ML113711, ML166177, ML168862, ML111711411, ML147607651, ML164703041, ML190304991, ML203944091, ML203957081, ML256858971, ML322721821, ML525798401, ML543249881, ML587528581, ML587529221, ML602096671, ML609065062, ML609065063, ML609065064, ML609677276, ML609677316, ML613641225, ML616826975, ML619671486, ML620439231, ML621028210, ML621096567, ML621096574, ML621310704, ML621372291, ML623281377, ML623309467, ML623309468, ML625742026, ML627956764, ML627956777, ML630573248, ML630573652, ML35357, ML56152, ML164605, ML164606, ML164608, ML164715, ML164716, ML164717, ML204519, ML26071721, ML87684741, ML88224041, ML179377811, ML179377851, ML203886221, ML232693731, ML358767001, ML436704181, ML561904951, ML561908881, ML561909951, ML562319841, ML566642741, ML566647111, ML566647171, ML566655341, ML566655451, ML566659011, ML602245231, ML609053734, ML609053868, ML611371661, ML612195775, ML616765940, ML624491795, ML624491962, ML624526629, ML624529467, ML632550939, ML165266, ML165268, ML243667, ML243668, ML243669, ML243670, ML243671, ML243672, ML243673, ML243674, ML243675, ML243676, ML245139, ML246286, ML246287, ML246288, ML246289, ML246290, ML246291, ML246292, ML246293, ML246294, ML267615, ML537448, ML537454, ML68154271, ML74640981, ML87785711, ML95373761, ML98961981, ML113513331, ML114075941, ML146126661, ML147457121, ML151907461, ML154146821, ML156518961, ML173503011, ML173504141, ML176634231, ML193068761, ML217077671, ML243610381, ML259934831, ML259934871, ML277785841, ML277786011, ML288929911, ML371599581, ML371599721, ML396109321, ML412938661, ML512840321, ML512861251, ML515394081, ML515394091, ML515394101, ML527409241, ML527409271, ML553386151, ML553386161, ML564910421, ML579470081, ML608638644, ML616392150, ML616410603, ML616552090, ML616552091, ML616552092, ML616552093, ML617118723, ML618524343, ML619062259, ML619747076, ML619747078, ML619747209, ML619881438, ML619881722, ML620130637, ML620798952, ML620799036, ML622774404, ML624219703, ML624369195, ML625693115, ML626148429, ML627169842, ML627651056, ML627672287, ML629314273, ML629314316, ML630316237, ML631059005, ML631059037, ML631059155, ML631096360, ML632590332, ML633001732, ML633002385, ML633002505, ML117978, ML139160, ML86087081, ML118003911, ML134612171, ML184332401, ML184332421, ML193247251, ML193247641, ML194869971, ML194874931, ML199716261, ML199716341, ML199817441, ML202245611, ML221587251, ML221587331, ML271568511, ML271568521, ML282613261, ML282613311, ML287702501, ML290031561, ML299740681, ML371394301, ML376706631, ML390438031, ML392836001, ML407902661, ML416539261, ML416541341, ML416543401, ML417025251, ML417402171, ML423108471, ML439563731, ML506063411, ML506064451, ML552991841, ML566493981, ML608662553, ML609239494, ML612072806, ML613773763, ML614142915, ML614145546, ML616082269, ML616082304, ML616082381, ML616607957, ML617774668, ML618579073, ML618962405, ML618962406, ML618962407, ML621973348, ML622721801, ML625038482, ML625546135, ML627855239, ML630949759, ML204733, ML204734, ML204735, ML204738, ML204741, ML204742, ML204744, ML204745, ML204747, ML204749, ML204751, ML204754, ML247725811, ML623217043, ML623217055, ML623217092, ML629692920, ML594742111, ML623103657, ML623103689.

\end{document}